\documentclass[journal]{IEEEtran}

\usepackage{cite}
\usepackage{graphicx}
\usepackage{amsmath}
\usepackage{mathtools}
\usepackage{amssymb}
\usepackage{makecell}
\usepackage{multirow}
\usepackage{bbm}
\usepackage{booktabs}
\usepackage[usenames,dvipsnames]{xcolor}
\graphicspath{{./images/}}
\usepackage{caption}
\usepackage{floatrow}
\usepackage{bbold}

\usepackage{soul}
\usepackage{color}
\usepackage{ulem} 
\usepackage{hyperref}

\usepackage{float}
\usepackage{booktabs}
\usepackage{multirow}
\usepackage{mathrsfs}
\usepackage[utf8]{inputenc}
\usepackage{subfigure}
\usepackage{threeparttable}
\usepackage{caption}
\usepackage{setspace}

\usepackage{pifont}

\definecolor{revisioncolor}{RGB}{0, 0, 0}  
\newcommand{\revised}[1]{\textcolor{revisioncolor}{#1}}
\definecolor{mypink1}{rgb}{0.858, 0.188, 0.478}
\definecolor{mygray}{gray}{0.6}

\newcommand{\colorcheck}{\textcolor{green!70!black}{\checkmark}}

\newcommand{\argmin}{\mathop{\rm arg~min}\limits}

\newcommand{\ie}{{\it i.e.}}
\newcommand{\eg}{{\it e.g.}}

\newcommand{\etc}{{\it etc.}}

\usepackage{rotating}
\usepackage{arydshln}

\usepackage{algorithm,algpseudocode}
\algnewcommand{\Inputs}[1]{%
\Statex \textbf{Inputs:}
  \Statex \hspace*{\algorithmicindent}\parbox[t]{.8\linewidth}{\raggedright #1}
}

\algnewcommand{\Outputs}[1]{%
\Statex\textbf{Outputs:}
  \Statex \hspace*{\algorithmicindent}\parbox[t]{.8\linewidth}{\raggedright #1}
}

\algnewcommand{\algorithmicforeach}{\textbf{for each}}
\algdef{SE}[FOR]{ForEach}{EndForEach}[1]
  {\algorithmicforeach\ #1\ \algorithmicdo}
  {\algorithmicend\ \algorithmicforeach}

\begin{document}
%

\title{\revised{Enhancing Multi-Camera Gymnast Tracking Through Domain Knowledge Integration}}
%
%
%

\author{Fan Yang, Shigeyuki Odashima, Shoichi Masui, Ikuo Kusajima, Sosuke Yamao, and Shan Jiang
        \thanks{
        Corresponding author: Fan Yang (fan.yang@fujitsu.com). }
        \thanks{
        All authors are with Fujitsu Research, Japan.} 
}

%



\maketitle


\begin{abstract}
  We present a robust multi-camera gymnast tracking, which has been applied at international gymnastics championships for gymnastics judging. Despite considerable progress in multi-camera tracking algorithms, tracking gymnasts presents unique challenges: (i) due to space restrictions, only a limited number of cameras can be installed in the gymnastics stadium; and (ii) due to variations in lighting, background, uniforms, and occlusions, multi-camera gymnast detection may fail in certain views and only provide valid detections from two opposing views. These factors complicate the accurate determination of a gymnast's 3D trajectory using conventional multi-camera triangulation. To alleviate this issue, we incorporate gymnastics domain knowledge into our tracking solution. Given that a gymnast's 3D center typically lies within a predefined vertical plane during \revised{much of their} performance, we can apply a ray-plane intersection to generate coplanar 3D trajectory candidates for opposing-view detections. More specifically, we propose a novel cascaded data association (DA) paradigm that employs triangulation to generate 3D trajectory candidates when cross-view detections are sufficient, and resort to the ray-plane intersection when they are insufficient. Consequently, coplanar candidates are used to compensate for uncertain trajectories, thereby minimizing tracking failures.
The robustness of our method is validated through extensive experimentation, demonstrating its superiority over existing methods in challenging scenarios. Furthermore, our gymnastics judging system, equipped with this tracking method, has been successfully applied to recent Gymnastics World Championships, earning significant recognition from the International Gymnastics Federation.
\end{abstract}


%
\IEEEpeerreviewmaketitle

\section{Introduction}\label{sec:intro}

\IEEEPARstart{G}{ymnastics} judging support systems (JSS) are designed to revolutionize the way gymnastics competitions are analyzed, offering efficient and professional analysis of gymnastics performance in both daily training and official events. By harnessing affordable hardware and advanced algorithms, we have successfully created a gymnastics JSS to accurately track and analyze the movements of gymnasts during their performance, providing near real-time feedback to make competitions more objective and exciting than ever before\footnote{\url{https://www.youtube.com/watch?v=CinAYBZYANg}}.

\begin{figure}[t]
	\centering	\includegraphics[width=\columnwidth]{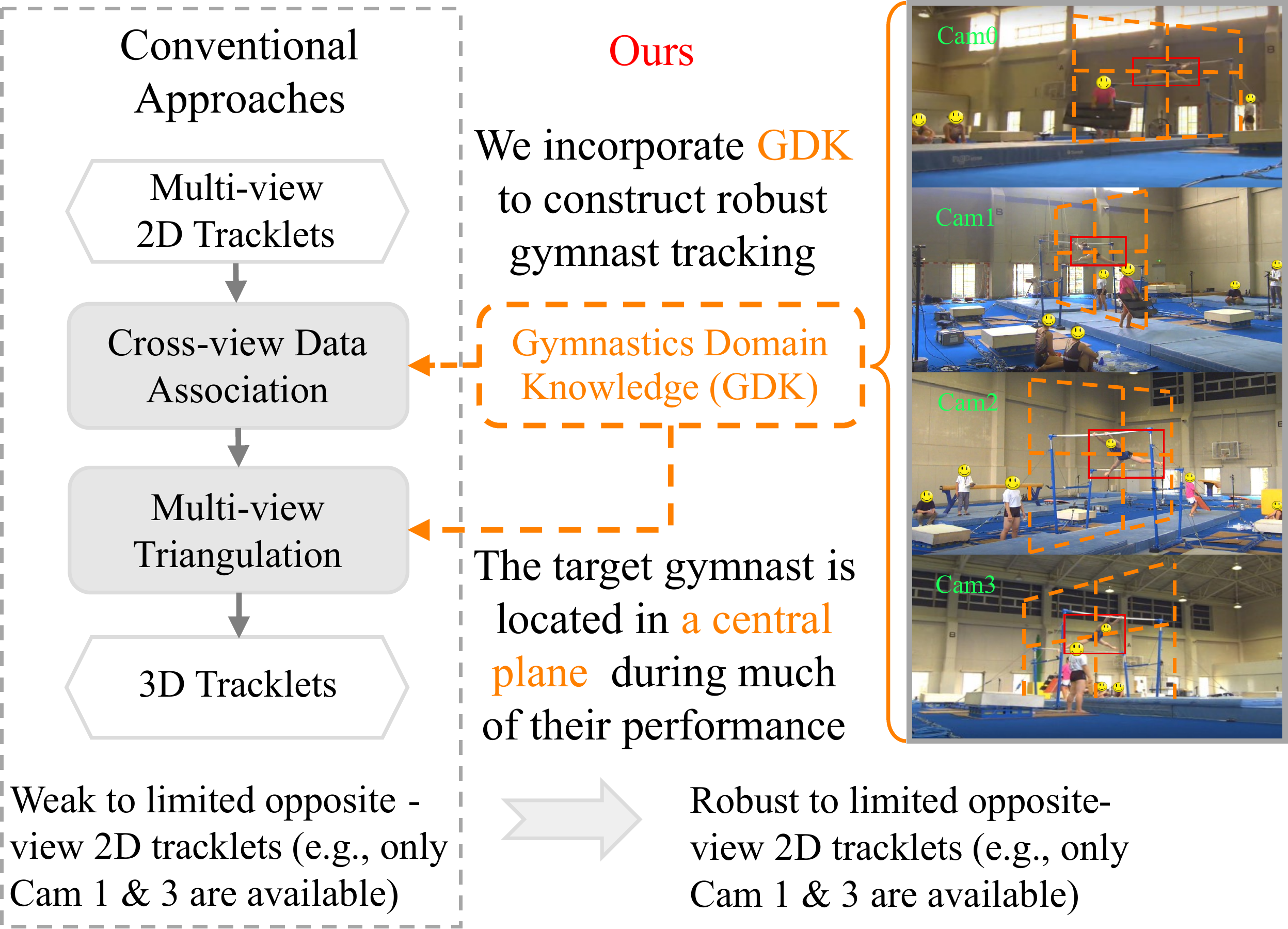}
	\caption{ \revised{\textbf{Overview of our proposal.} While conventional multi-camera multi-target tracking may fail due to limited opposite-view observations, our method incorporates gymnastics domain knowledge to improve tracking robustness.}}
	\label{fig:overview}
\end{figure}

The successful deployment of our judging system relies on robust gymnast tracking. As the gymnastics stadium could be crowded, before judging the performance, we need to detect and track the target gymnast among a large number of non-target spectators, team members, and coaches. Considering that gymnasts wear distinct uniforms across teams and similar uniforms within the same team, training a gymnast detector to directly detect the target gymnast is infeasible. In fact, gymnasts are treated as targets during their performance, and become non-targets once they exit the performance area. Therefore, we employ multi-camera tracking to obtain 3D trajectories to identify target gymnasts.

To make a trade-off between the system performance and available space in the gymnastics stadium, our tracking system takes four calibrated RGB cameras to capture the performing area from diverse viewpoints. Given four RGB video streams, parallelized person detectors (\eg, YOLOX-S~\cite{yolox2021}) are employed to generate bounding boxes (bboxes) on each camera view. These multi-view bboxes serve as inputs for our tracking framework, which produces 3D tracklets and corresponding multi-view 2D tracklets. We identify the target gymnast using 3D tracklets and perform multi-view 3D pose estimation using corresponding multi-view 2D tracklets. As shown in Fig.~\ref{fig:overview}, our system adopts a top-down approach~\cite{moon2019camera} and exclusively selects the target gymnast's tracklets to generate 3D poses, enabling higher-resolution pose estimation on targets and eliminating unnecessary computation on non-targets individuals.

\begin{table*}[ht!]
	\footnotesize
	\centering
	\caption{\revised{\textbf{Comparison of related works.} Previous methods of gymnast tracking primarily concentrated on 2D Single-Camera MOT, which could fail to generate accurate 3D poses in the downstream process. Other existing studies on Multi-Camera MOT assume the availability of ample cross-view detections for above-ground target tracking. Our study is the first to integrate domain knowledge to generate robust 3D tracklets from limited opposite-view matches.}}
	\label{tab:comparison}
	\begin{tabular}{lcccc}
	\textbf{MOT Studies} & \textbf{\makecell[l]{Multi-camera\\Setting}} & \textbf{\makecell[l]{ Target Bottom 3D\\ Height $>$ 0 meter }} & \textbf{\makecell[l]{Resilient 3D Tracklets for\\ Opposite-View Matches}} & \textbf{\makecell[l]{Integrate Domain\\Knowledge}} \\
	\toprule
	SC-MOT for Gymnastics~\cite{shao2020finegym} &  &   &  &  \\
	SC-MOT for Other Sports~\cite{cioppa2022soccernet, cui2023sportsmot, scott2024teamtrack} &  &   &  &  \\
	\makecell[l]{MC-MOT w/ Target Bottom On Ground\\ \cite{fleuret2007multicamera, yoo2016online,chen2016equalized,wen2017multi,he2020multi, he2020multi,nguyen2022lmgp,huang2023enhancing}} & \checkmark &   &  &  \\
	\makecell[l]{MC-MOT w/ Target Bottom Above Ground \\ 
 \cite{chen2020multi,chen2020cross,ohashi20vmocap,dong2021fast,yang2023unified}} & \checkmark & \checkmark  &  &  \\
	\textbf{Ours} & \checkmark & \checkmark & \colorcheck & \colorcheck \\
	\bottomrule
	\end{tabular}
\end{table*}

To accommodate gymnastics tracking, our implementation differs from existing multi-camera trackers. While a group of studies~\cite{wen2017multi, he2020multi, nguyen2022lmgp, huang2023enhancing} focus on tracking the footprints of individuals on flat ground with a constant vertical height, our framework addresses the unique challenge of tracking gymnasts who may jump to different vertical heights, which introduces additional complexity to the cross-camera data association (DA).

\revised{To effectively track gymnastic movements, our methodology diverges from conventional multi-camera tracking systems. While a subset of studies~\cite{wen2017multi, he2020multi, nguyen2022lmgp, huang2023enhancing} primarily track targets at a uniform elevation (\eg, on a flat ground), our tracking target---the gymnast---may jump to variable vertical heights. This variability introduces significant complexity to the cross-camera data association (DA) process. Additionally, other research~\cite{chen2020multi, chen2020cross, dong2021fast, yang2023unified} has demonstrated the ability to track targets over varying vertical heights with adequate multi-view detections. Nevertheless, due to variations in lighting, background, uniforms, and occlusions when capturing gymnastics, we may have only opposite-view detections available for multiple successive frames. }
In this difficult case, a small bias in the 2D detections can be amplified to unacceptable shifting errors in the corresponding 3D positions. 

\revised{It is important to recognize that these shifting errors in 3D tracklets are not indicative of deficiencies in the data association of tracking, but rather are intrinsic consequences of the 3D triangulation process when opposite-view matches are provided~\cite{andrew2001multiple}. The fundamental properties of 3D triangulation inherently impose limitations on the accuracy of existing multi-camera tracking methods, irrespective of their state-of-the-art (SOTA) performance. Even when these methods accurately match perfect detections from opposing views, spatial displacement in the resulting 3D tracklets is inevitable.}

We alleviate this issue by utilizing gymnastics domain knowledge: gymnasts often lie in predefined vertical planes during \revised{much of their} performances. Therefore, we utilize triangulation to generate 3D trajectory candidates when detections are sufficient and switch to ray-plane intersection otherwise. Valid candidates are then merged under a cascade tracking paradigm. In this way, gymnastic domain knowledge is leveraged to compensate for uncertain trajectories and reduce tracking failures.

\revised{For the first time in this research domain, we have collected multi-camera gymnastics data in authentic gymnastics studios to conduct innovative experiments. Our dataset encompasses a variety of gymnastic disciplines, including Balance Beam (BB), Pommel Horse (PH), Vault (VT), Still Rings (SR), Horizontal Bar (HB), Uneven Bars (UB), and Parallel Bars (PB). We validate our tracking technology by comparing our results with the annotated ground truth, confirming its robustness in challenging scenarios where only opposite-view detections are available for multiple successive frames. Benefiting from these robust tracking results, we are able to achieve accurate gymnastic judging performance. Our judging system, equipped with this advanced tracking technology, has been successfully applied at the Gymnastics World Championships and has garnered significant recognition. This success highlights the effectiveness and practicality of our approach}

In summary, our contributions are two-fold: 
\begin{itemize}
    \item We presented a compelling case study that introduced domain knowledge to improve the robustness of multi-camera tracking. Compared to existing works, it can significantly reduce tracking failures in challenging scenarios with sparse multi-view detections, providing valuable insights for designing robust domain-specific tracking in other related applications.
	\item Our multi-camera gymnast tracking framework has been integrated into the gymnastics judging system and is now in active use during daily training sessions and international championships. This has revolutionized the field of gymnastics video processing, \revised{providing innovative applications not only enhance athlete performance but also improve both the efficiency and impartiality of gymnastic judgments}.
\end{itemize}


\section{Related Works}\label{sec:related}

\revised{Initially, we compare our proposed method with related works across several dimensions, and subsequently, we outline our principal distinctions and innovations in TABLE~\ref{tab:comparison}.}

\subsection{Multi-object Tracking (MOT) in Sport Video Analysis} 

The dynamic and often chaotic nature of sports presents a unique challenge for video analysis. Precisely tracking the movements and interactions of athletes amidst rapid changes in scene and perspective is crucial for gleaning valuable insights from sports footage. 
Multi-Object Tracking (MOT) has emerged as a pivotal tool in sports video analysis, facilitating the automated identification and localization of target objects over time~\cite{yu2017adaptive, sheng2018iterative, sheng2018heterogeneous, xiang2020end, you2020multi, luna2022graph}. The  resulting outputs---tracklets---provide valuable insights into player movements, strategies, and overall game progression~\cite{li2009automatic, shih2017survey, kong2019joint, wu2019ontology, qi2019sports}. This, in turn, contributes to enhanced performance analysis, tactical planning, and injury prevention. Therefore, the development and refinement of robust and efficient MOT algorithms are of paramount importance in advancing the field of sports video analysis.

In a pioneering effort to facilitate high-performance gymnast tracking for assisting judging in professional gymnastics championships, we have uniquely integrated domain-specific knowledge of gymnastics into the design of a novel and robust MOT framework. This innovative approach marks a significant advancement in the application of MOT within the field of professional gymnastics.

\subsection{Single-camera Multi-object Tracking (SC-MOT)} 

Among various settings of MOT, SC-MOT techniques rely on monocular camera inputs to track multiple objects~\cite{DeepSORT, 1517Bochinski2017, zhang2022bytetrack, yang2023hard}. These approaches include appearance-based methods, motion-based methods, and hybrid methods that combine both appearance and motion cues. Nonetheless, recovering our target 3D tracklet from a single-camera RGB video is an ill-pose problem, except for tracking the 3D footprints on flat ground with a constant vertical height~\cite{wen2017multi, he2020multi, you2020multi, nguyen2022lmgp, huang2023enhancing, du2023StrongSORT}. Other related works~\cite{luiten2020track, kim2021eagermot,  wang2022deepfusionmot} apply single depth camera to obtain 3D tracklets with varying heights, but depth sensors may obtain sparse captures at far distances and long-term operation risks distortions caused by hardware overheating. To circumvent these problems, we opt for using multiple RGB cameras. However, the SC-MOT technique is beneficial for multi-camera tracking. As presented in UniMMT~\cite{yang2023unified}, using the motion-based SC-MOT method to construct single-camera 2D tracklets, and subsequently associating 2D tracklets to 3D tracklets by cross-camera triangulation, leads to an enhancement in the performance of multi-camera tracking.

\subsection{Multi-camera Multi-object Tracking (MC-MOT)} 

Existing MC-MOT approaches can be categorized into two main groups. The first focuses on tracking 3D footprints on the ground plane with a constant vertical height~\cite{fleuret2007multicamera, yoo2016online, chen2016equalized, wen2017multi, chen2020multi,he2020multi, nguyen2022lmgp, luna2022graph, huang2023enhancing}. This is achieved by projecting 2D bottom points onto the ground plane using homography matrices, thereby generating 3D footprints from various views without the need for explicit triangulation. However, as our gymnast tracking involves varying jump heights, these methods are not applicable. 
The second category explores 3D pose tracking with diverse vertical heights~\cite{chen2020multi,chen2020cross,dong2021fast, yang2023unified}. 
Among them, UniMMT~\cite{yang2023unified} effectively addresses 2D-3D ambiguities in multi-camera tracking by jointly leveraging spatiotemporal constraints for cross-frame cross-view data association (DA). Our gymnast tracking framework adopts UniMMT's basic components, while also incorporating gymnastics-specific knowledge to enhance robustness.

\subsection{Cascaded Data Association in MOT}

Data association (DA) serves as a potent mechanism  for establishing connections of cross-frame and cross-source detections in MOT~\cite{ciaparrone2020deep, sun2020survey, luo2021multiple}. In SC-MOT works, DA is employed to link cross-frame detections. Typically, the distance matrix of cross-frame detections is calculated and the Hungarian algorithm~\cite{kuhn1955hungarian} is applied to obtain the cross-frame matching. However, DA's utility extends beyond linking cross-frame detections in SC-MOT studies. It is also leveraged for cross-camera matching. Given that the 3D tracklet of an identical object may have multiple 2D projected tracklets in different camera views, graph-based clustering methods are generally employed to match and fuse cross-view tracklets~\cite{fleuret2007multicamera, yoo2016online, chen2016equalized, wen2017multi, chen2020multi,he2020multi, nguyen2022lmgp, luna2022graph, huang2023enhancing}. 

To better manage changes in the appearance, occlusions, and noise of targets objects, cascaded DA is further developed to prioritize the processing of confident and straightforward detections, followed by the processing of more ambiguous and difficult detections. For instance, DeepSORT~\cite{DeepSORT} associates new detections with recently matched tracks over earlier matched tracks; ByteTrack~\cite{zhang2022bytetrack} matches confident detections earlier than ambiguous ones; C-BIoU~\cite{yang2023hard} associates alive tracks and detections with a small buffer before a large buffer; EagerMOT~\cite{kim2021eagermot} utilizes 2D detections to support tracking after 3D DA fails; and DeepFusionMOT~\cite{wang2022deepfusionmot} applies more complicated cascaded DA to associate camera-LiDAR tracks.

Inspired by these studies, we employ the domain knowledge of gymnastics to form a novel cascaded DA scheme. To associate cross-camera 2D tracklets to 3D trajectories, we prioritize the use of cross-camera triangulation when the detections are adequate, and alternatively employ the ray-plane intersection when necessary (see Fig.~\ref{fig:framework}).


\section{Method}\label{sec:method}

\begin{figure*}[th!]
	\begin{center}
	\hsize=\textwidth
	\includegraphics[width=\textwidth]{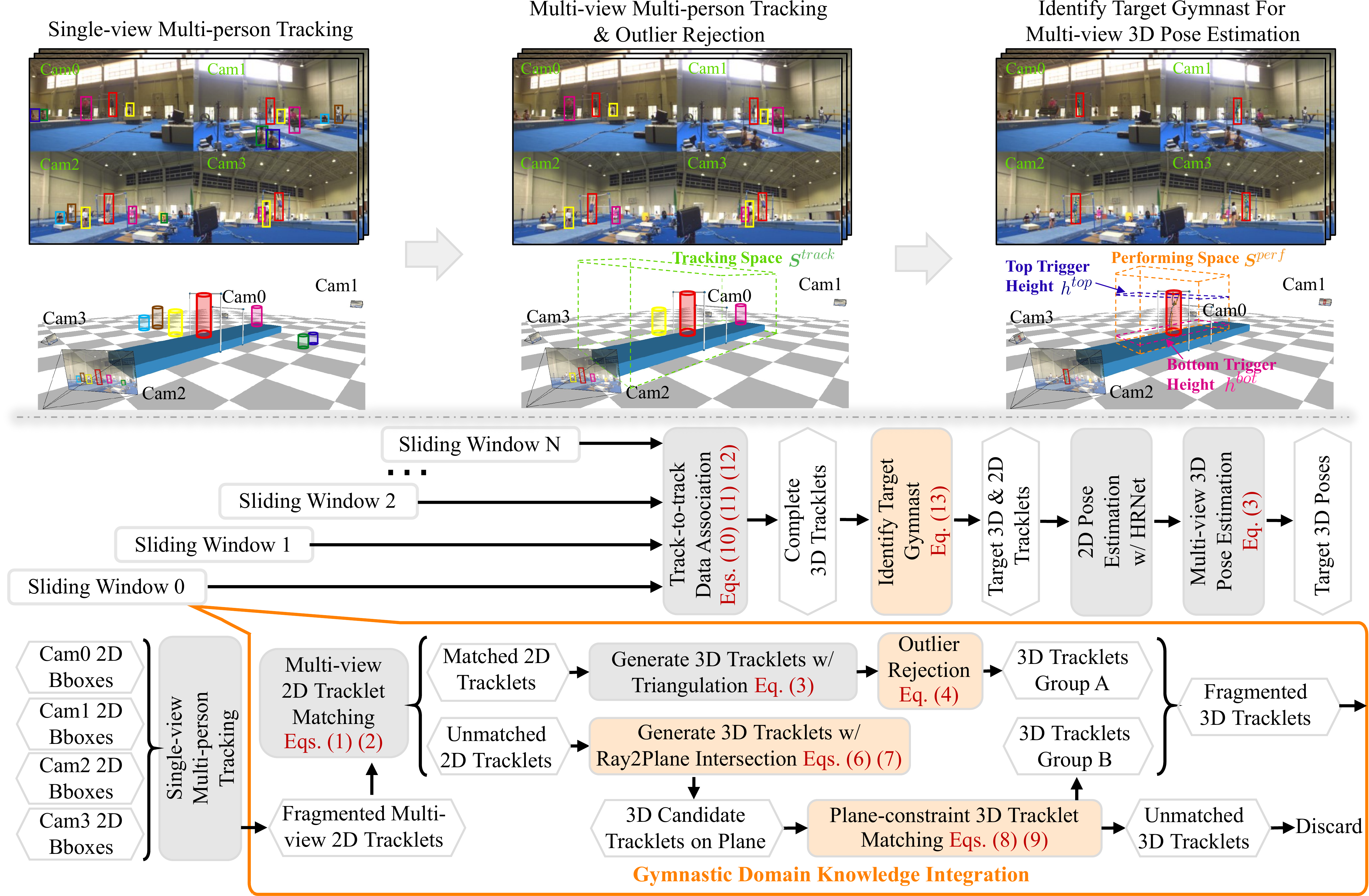}
	\caption{\textbf{Architecture of our gymnast tracking framework.} We segment 2D tracklets into fragments in real time and collaboratively refine them using multi-view information. By incorporating gymnastics domain knowledge, we apply triangulation to generate 3D tracklet candidates when detections are sufficient and resort to ray-plane intersection when they are not. Valid candidates are subsequently fused under a cascaded data association paradigm. Thereafter, we reassemble fragmented tracklets by integrating both 2D and 3D information and identify the tracklet corresponding to the target gymnast. Utilizing the associated multi-view 2D tracklets of the target gymnast, we generate 3D poses and employ them for judging purposes. Note that, while the {\color{lightgray} gray-colored modules} are partially derived from the precursor work, UniMMT~\cite{yang2023unified}, the \textcolor{orange}{orange-colored modules} have been newly developed specifically for gymnast tracking.}
\label{fig:framework}
\end{center}
\end{figure*}

The architecture of our method is illustrated in Fig.~\ref{fig:framework}.  
Our tracking framework utilizes synchronized multi-camera 2D bounding boxes as inputs and produces the 3D tracklet and corresponding multi-camera 2D tracklets for the target gymnast as outputs. We further generate the 3D poses according to multi-camera 2D tracklets. As we focus on the tracking framework, we omit the details of camera calibration, model training for gymnast detection and pose estimation, and pose-based judging, despite their importance for our gymnastics judging system.

\subsection{Single-view Processing}

At frame $t$, a bbox with index $i$ from camera view $c$ can be represented as $\boldsymbol{b}_{t,c,i} = [x_{t,c,i}, y_{t,c,i}, w_{t,c,i}, h_{t,c,i}]$, which includes the center-point coordinate, the width, and the height, respectively. Initially, we employ parallelized IoU trackers~\cite{1517Bochinski2017} to obtain 2D tracklets for each camera view. Each camera view can have multiple 2D tracklets. From frame $t_0$ to $t_n$, a complete tracklet is denoted as $\mathcal{T}^{2D}_{t_0:t_n,c,k} = \{\boldsymbol{b}_{t_0,c,0}, \dots, \boldsymbol{b}_{t_n,c,n}\}$, where $k$ represents the tracklet index.

Compared to trackers that integrate both appearance and motion cues (\eg, DeepSORT~\cite{DeepSORT}), the IoU tracker solely relies on the motion cue. While this allows for a faster speed ($>10,000$ FPS), it may lead to an increased number of identity (ID) switches~\cite{1517Bochinski2017}. To address this issue, we follow UniMMT~\cite{yang2023unified} to segment imperfect 2D tracklets into sections and collaboratively refine them by incorporating both 2D and 3D information.

In \revised{detail}, a complete 2D tracklet can be divided into multiple temporally overlapping segments, denoted as $\big\{\mathcal{T}^{2D}_{t_0:t_0+\omega ,c,k}, \mathcal{T}^{2D}_{t_0+0.5\omega :t_0+1.5\omega ,c,k}, \dots, \mathcal{T}^{2D}_{t_n-\omega :t_n,c,k}\big\}$, where $\omega $ represents the temporal sliding window range and the step size is $0.5~\omega $. Since the movement of our target gymnast is fast and our video recording speed is $30$ FPS, we experimentally set $\omega =10$ frames. In a practical setting, while multiple IoU trackers operate parallelly in an online manner, we concurrently extract segments in real-time every $\omega $ frames. Within a sliding window, we interpolate or extrapolate missing frames if there are at least $5$ elements and discard the others.

\subsection{Multi-view Processing}

After obtaining fragmented multi-view 2D tracklets, we utilize their cross-camera geometric relationship to match them. First, let us denote the intrinsic and extrinsic parameters of camera $c$ as $\mathbf{K}_c \in \mathbb{R}^{3 \times 3}$ and $[\mathbf{R}_c \in \mathbb{R}^{3 \times 3} |\mathbf{t}_c \in \mathbb{R}^{3 \times 1}]$, respectively. The corresponding projection matrix can be represented as $\mathbf{P}_{c} = \mathbf{K}_c [\mathbf{R}_c|\mathbf{t}_c] \in \mathbb{R}^{3 \times 4}$. The fundamental matrix between camera $c_i$ and camera $c_j$ is denoted as $\mathbf{F}_{c_i,c_j} \in \mathbb{R}^{3 \times 3}$~\cite{andrew2001multiple}. 
Next, we calculate the cross-view distance $\mathrm{d}^{x}$ between two bboxes $\boldsymbol{b}_{t,c_i,i}$ and $\boldsymbol{b}_{t,c_j,j}$, which can be formulated as
\begin{equation}
\resizebox{\columnwidth}{!}{$
\begin{split} 
	\mathrm{d}^{x} \left (\boldsymbol{b}_{t,c_i,i}, \boldsymbol{b}_{t,c_j,j} \right ) =& \frac{\mathrm{d} \big( \left [x_{t,c_i,i}, y_{t,c_i,i}\right ], \boldsymbol{l}_{c_{i}}\left ( \left [x_{t,c_j,j}, y_{t,c_j,j}\right ] \right ) \big)}{|w_{t,c_i,i}+h_{t,c_i,i}|} \\
		+& \frac{\mathrm{d} \big(\left[x_{t,c_j,j}, y_{t,c_j,j}\right], \boldsymbol{l}_{c_{j}}\left (\left [x_{t,c_i,i}, y_{t,c_i,i}\right ] \right ) \big) }{|w_{t,c_j,j}+h_{t,c_j,j}|},  \\
		\textrm{with}~~~~~~~~~~~~~~~~~~~~~~ \\
		\boldsymbol{l}_{c_{i}}\left (\left[x_{t,c_j,j}, y_{t,c_j,j}\right]\right ) =& \bold{F}_{c_{j},c_{i}}\left [x_{t,c_j,j}, y_{t,c_j,j}, 1 \right ],\\
		\boldsymbol{l}_{c_{j}}\left (\left[x_{t,c_i,i}, y_{t,c_i,i}\right]\right ) =& \bold{F}_{c_{i},c_{j}}\left [x_{t,c_i,i}, y_{t,c_i,i}, 1 \right ],\\
	\boldsymbol{b}_{t,c_i,i} =& \left [x_{t,c_i,i}, y_{t,c_i,i}, w_{t,c_i,i}, h_{t,c_i,i} \right ],\\
	\boldsymbol{b}_{t,c_j,j} =& \left [x_{t,c_j,j}, y_{t,c_j,j}, w_{t,c_j,j}, h_{t,c_j,j}\right ],
\end{split} 
\label{eq:D_epipolar_dist}
$}
\end{equation}
where the epipolar line $\boldsymbol{l}$ can be represented by three parameters as $(l_1, l_2, l_3)$ and the point-to-line distance $\mathrm{d} \left ([x,y], \boldsymbol{l} \right )= \frac{|l_1 x + l_2 y + l_3|}{\sqrt{l_1^2 + l_2^2}}$. We normalize it by the bbox scale $|w+h|$.

Within a sliding window defined by the interval $t_s:t_s+\omega $, we calculate the cross-view distance between the 2D tracklets $\mathcal{T}^{2D}_{t_s:t_s+\omega ,c_i,k_i}$ and $\mathcal{T}^{2D}_{t_s:t_s+\omega ,c_j,k_j}$ as follows:
\begin{equation}
\resizebox{\columnwidth}{!}{$	
\begin{split}  
	\mathrm{d}^{x}(\mathcal{T}^{2D}_{t_s:t_s+\omega ,c_i,k_i}, \mathcal{T}^{2D}_{t_s:t_s+\omega ,c_j,k_j}) =~~~~~~~~~~~~~~~~~~~~~~~~~~~~~~~~~~~~~~~\\
	\left\{\begin{matrix}
	& \textrm{mean} \big ( \left\{\mathrm{d}^{x} \left (\boldsymbol{b}_{t,c_i,i}, \boldsymbol{b}_{t,c_j,j} \right ) \right\} \big ), & \text{if} & \Psi_{i} \cap  \Psi_{j} \neq \emptyset ~\text{and}~ c_{i} \neq c_{j},\\ 
	& \emptyset, & \text{if} & \Psi_{i} \cap  \Psi_{j} =  \emptyset ,\\
	& \inf, &\text{if} & \Psi_{i} \cap  \Psi_{j} \neq \emptyset~\text{and}~c_{i} = c_{j} ,
	\end{matrix} \right.\\
	\forall ~~\boldsymbol{b}_{t,c_i,i} \in \mathcal{T}^{2D}_{t_s:t_s+\omega ,c_i,k_i},~ 
	\boldsymbol{b}_{t,c_j,j} \in \mathcal{T}^{2D}_{t_s:t_s+\omega ,c_j,k_j},~~~~~~~~~~~
\end{split}
\label{eq:2d_tracklets_dist}
$}
\end{equation}
where $\Psi_{i} \in \{t_s:t_s+\omega \}$ and $\Psi_{j} \in \{t_s:t_s+\omega \}$ respectively denote the valid frame sets of two cross-view 2D tracklets.

To handle disjointed-frame tracklets (\ie, $\Psi_{i} \cap  \Psi_{j} =  \emptyset$) in our cross-view 2D tracklets clustering, we have adapted the distance updating function of the complete-linkage clustering algorithm~\cite{murtagh2012algorithms} to suit our specific needs. This modification is detailed in Algorithm~\ref{algorithm}. The inputs for this algorithm are the cutting thread, denoted as $\lambda$, and the cross-view 2D tracklets $\big\{\mathcal{T}^{2D}_{t_s:t_s+\omega ,c_j} \mid c_i \in \{0,1,2,3\},~k_i \in \{0,\dots,k_n\} \big\}$, which are within the same sliding window. We calculate distances between all cross-view 2D tracklets and put them into a set $\big\{\mathrm{d}^{x}(\mathcal{T}^{2D}_{t_s:t_s+\omega ,c_i,k_i}, \mathcal{T}^{2D}_{t_s:t_s+\omega ,c_j,k_j}) \mid  c_i,c_j \in \{0,1,2,3\}~\textrm{and}~k_i,k_j \in \{0,\dots,k_n\} \big\}$. Referring to this set, multi-view 2D tracklets corresponding to the identical individual are associated into a unique cluster. Subsequently, 3D positions can be inferred from the multi-view 2D positions. 

\begin{algorithm} 
	\caption{Our Cross-camera Clustering}
	\label{algorithm}
	\begin{algorithmic}[1]
	\State Initialize a cluster set $\mathbf{\Theta}$ with the size $N_{\mathbf{\Theta}}$, where each $\mathbf{\Theta}[i]$ contains a single 2D tracklet $\mathcal{T}^{2D}$.
	\State Initialize a distance matrix $\mathbf{D}[i,j] = \mathrm{d}^{x}(\mathbf{\Theta}[i], \mathbf{\Theta}[j] )$ by Eq.~\ref{eq:2d_tracklets_dist}.
	\While{$N_{\mathbf{\Theta}}>1$ and $\exists~ \mathbf{D}[i,j]< \lambda$ }
		\State Find the closest cluster pair $\left \langle  \mathbf{\Theta}[i], \mathbf{\Theta}[j] \right \rangle \gets \argmin_{i,j \in \{1, \dots, N_{\mathbf{\Theta}}\}} \mathbf{D}$
		\State Merge $\left \langle  \mathbf{\Theta}[i], \mathbf{\Theta}[j] \right \rangle$ to a new cluster $\mathbf{\Theta}[k] \gets \mathbf{\Theta}[i] \cup \mathbf{\Theta}[j]$
		\State Update the cluster set $\mathbf{\Theta} \gets \mathbf{\Theta}\setminus \{\mathbf{\Theta}[a], \mathbf{\Theta}[b]\} \cup \{\mathbf{\Theta}[k]\}$
		\For{$q \in \{1, \dots, N_{\mathbf{\Theta}}\} \setminus \{k\}$} \Comment{{\color{SkyBlue}{Update $\mathbf{D}$}}}
		\Statex \Comment{{\color{SkyBlue}{Traditional complete-linkage clustering approach:~~~~~~~}}}
		\State \textcolor{mygray}{\sout{$\mathbf{D}[k,q] \gets \text{max} \left ( \mathbf{D}[i,q], \mathbf{D}[j,q] \right )$}} 
		\Statex \Comment{{\color{SkyBlue}{Our approach:~~~~~~~~~~~~~~~~~~~~~~~~~~~~~~~~~~~~~~~~~~~~~~~~}}}
		\If{$\mathbf{D}[i,q] \neq  \emptyset ~\text{and}~ \mathbf{D}[j,q] \neq  \emptyset$}
		\State $\mathbf{D}[k,q] \gets \text{max} \left ( \mathbf{D}[i,q], \mathbf{D}[j,q] \right )$
		\ElsIf{$\mathbf{D}[i,q] =  \emptyset ~\text{and}~ \mathbf{D}[j,q] \neq  \emptyset$}
		\State $\mathbf{D}[k,q] \gets  \mathbf{D}[j,q]$
		\ElsIf{$\mathbf{D}[i,q] \neq \emptyset ~\text{and}~ \mathbf{D}[j,q] =  \emptyset$}
		\State $\mathbf{D}[k,q] \gets  \mathbf{D}[i,q]$
		\ElsIf{$\mathbf{D}[i,q] = \emptyset ~\text{and}~ \mathbf{D}[j,q] =  \emptyset$}
		\State $\mathbf{D}[k,q] \gets  \emptyset$
		\EndIf
	  \EndFor
	\EndWhile
\end{algorithmic}
\end{algorithm}

Given a valid cluster of matched cross-view 2D tracklets, we represent the set of matched camera views and 2D positions by $\mathcal{C} \in \{0,1,2,3\}$ and $\left \{\mathbf{x}_{t,c_i,i} = \left [x_{i},y_{i},1 \right ] \mid c_i \in \mathcal{C} \right \}$, respectively. The corresponding 3D position $\mathbf{X}_{t,k_i} \in \mathcal{T}^{3D}_{t_s:t_s+\omega , k_i}$ can be estimated by solving the following equation~\cite{andrew2001multiple}:
\begin{equation}\label{eq:3d_triangulation}
\begin{split}
	\mathbf{X}_{t,k_i} = \arg\min_{\mathbf{X}_{t,k_i}} \sum_{\forall c_i \in \mathcal{C}}\left\| \mathbf{x}_{t,c_i,i} - \pi \left (\mathbf{P}_{c_i}, \mathbf{X}_{t,k_i} \right ) \right\|^2
\end{split}
\end{equation}
where $\mathbf{P}_{c_i}$ is the 3D-to-2D projection matrix from camera $c_i$ and $\pi(\cdot)$ represents the projection function that maps 3D points to 2D points. This formula can be employed to generate 3D tracklets for all associated multi-view 2D tracklets.

In a crowded scene, we may obtain more 3D tracklets than we need. As shown in Fig.~\ref{fig:framework}, gymnastics \revised{movement} is restricted within a predefined 3D performance space, represented as a cuboid and denoted by $\boldsymbol{S}^{perf} = [x_{min},y_{min},z_{min}, x_{max},y_{max},z_{max}]$~\cite{de20172020, de2018technical}. To account for potential variations, we add a spatial buffer $\beta$ to this space, thereby defining a tracking space, which is denoted as $\boldsymbol{S}^{track} = [x_{min}-\beta,y_{min}-\beta,z_{min}, x_{max}+\beta,y_{max}+\beta,z_{max}]$. Given a 3D tracklet $\mathcal{T}^{3D}_{t_s:t_s+\omega , k_i}$, it is classified as an outlier if it is located outside the tracking space or exhibits abnormal velocity, which can be expressed as follows
\begin{equation}\label{eq:outlier}
\begin{split}
	\textrm{outlier} \left (\mathcal{T}^{3D}_{t_s:t_s+\omega , k_i}, \boldsymbol{S}^{track}, \nu \right ) =~~~~~~~~~~~~~~~~~~~~~~\\
	\begin{cases}
		1, & \text{if}~\mathbf{X}_{t,k_i} \notin \boldsymbol{S}^{track}~\text{or}~ \left \|\mathbf{X}_{t+1,k_i} - \mathbf{X}_{t,k_i}  \right\|_2 > \nu  \\
		0, & \text{otherwise},
	\end{cases} \\
	\forall  ~\mathbf{X}_{t,k_i} \in   \mathcal{T}^{3D}_{t_s:t_s+\omega , k_i}, ~ t \in  [t_s:t_s+\omega], ~~~~~~~~~~~~~
\end{split}
\end{equation}
where $\nu$ is the velocity threshold. Since our capture frame rate is $30$ FPS, we experimentally set $\nu=1$ meter.
We remove detected outliers to eliminate unnecessary downstream computations.

\begin{figure}[th!]
	\centering	\includegraphics[width=\columnwidth]{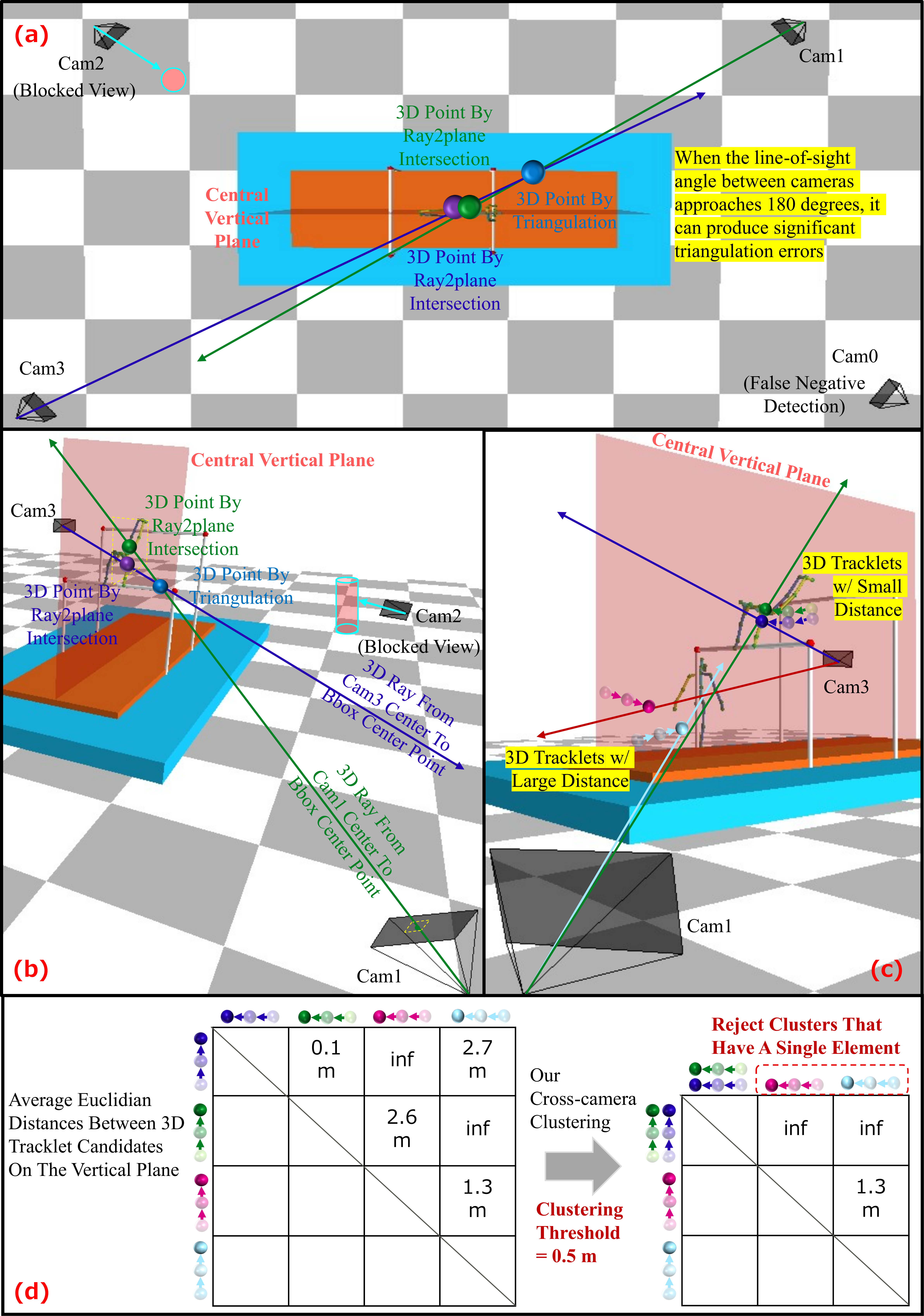}%
	\caption{\textbf{3D position estimation.} (a) and (b) present a challenging scenario in which only two opposing-view detections are available. While (a) shows the top-view scene, (b) is the corresponding side-view scene. Both (a) and (b) demonstrate that using our ray-to-plane intersection yields better 3D positions compared to using conventional triangulation. (c) illustrates that, while coplanar 3D tracklets are generated for multiple persons, coplanar 3D tracklets of the target gymnast generally have small gaps, allowing them to be grouped together. (d) depicts how 3D tracklet candidates of (c) are matched on the vertical plane. }
	\label{fig:ray2plane}
\end{figure}

Up to this point, our tracking components may be able to generate satisfactory tracking results when high-quality detections are available on more than three camera views~\cite{yang2023unified}. However, as shown in Fig.~\ref{fig:ray2plane} (a) and (b), detection failures may result in a challenging situation where only two-view detections are available, and the line-of-sight angle between these two views is nearly $180$ degrees. In such cases, a minor bias of the detections may lead to substantial errors in the 3D positions.
To alleviate this issue, we introduce domain knowledge from gymnastics, as the target gymnast typically moves within a predefined vertical plane (\ie, the \textcolor{pink}{pink plane} in Fig.~\ref{fig:ray2plane}). Unlike the triangulation method (\ie, Eq.~(\ref{eq:3d_triangulation})), which requires at least two cross-view 2D positions for 3D position estimation, the ray-to-plane intersection allows us to derive a 3D position on the target plane using a single 2D position~\cite{andrew2001multiple}.

We propose a novel cascaded DA to effectively incorporate triangulation and ray-to-plane intersection. As illustrated in Fig.~\ref{fig:framework}, we examine the multi-view association results: if an associated cluster only includes a single 2D tracklet or two 2D tracklets from opposite views, these samples are considered as unmatched 2D tracklets. Consequently, they are not subjected to triangulation processing. Instead, we use ray-plane intersection to generate 3D tracklet candidates from these samples, then perform additional matching with our cascaded DA.

In unmatched 2D tracklets, for a bbox with a center on image coordinate $[x_{t,c,i}, y_{t,c,i}]$, we can form a vector from the camera center $[0,0,0]$ to this image point in the camera coordinate system by
\begin{equation}\label{eq:VecCam}
	{}^{cam}\mathbf{v}_{t,c,i} = [\frac{x_{t,c,i}-\mathbf{K}_c[0,2]}{\mathbf{K}_c[0,0]}, \frac{y_{t,c,i}-\mathbf{K}_c[1,2]}{\mathbf{K}_c[1,1]}, 1],
\end{equation}
where $\mathbf{K}_c[0,0]$ and $\mathbf{K}_c[1,1]$ are camera focal lengths, $\mathbf{K}_c[0,2]$ and $\mathbf{K}_c[1,2]$ are principal point offsets. We further transform this vector from camera coordinate to world coordinate with
	\begin{equation}\label{eq:VecWorld}
	{}^{w}\mathbf{v}_{t,c,i} = [\frac{x_{t,c,i}-\mathbf{K}_c[0,2]}{\mathbf{K}_c[0,0]}, \frac{y_{t,c,i}-\mathbf{K}_c[1,2]}{\mathbf{K}_c[1,1]}, 1] \mathbf{R}_c .
\end{equation}

In the world coordinate system, a vertical plane can be defined with a normal vector $\mathbf{n}$ and a point $\mathbf{X^p}$ on the plane, while the center of camera $c$ is $\mathbf{X^o_c} = -\mathbf{R_c}^\intercal \mathbf{t_c}$. We calculate the intersection point $\mathbf{X}_{t,c,i}$ as follows:
\begin{equation}\label{eq:intersection}
	\mathbf{X}_{t,c,i} = \mathbf{X^o_c} + \frac{\mathbf{n} \cdot (\mathbf{X^p} - \mathbf{X^o_c}) }{\mathbf{n} \cdot {}^{w}\mathbf{v}_{t,c,i} } ~{}^{w}\mathbf{v}_{t,c,i},~~\forall~\mathbf{X}_{t,c,i} \in \mathcal{T}^{3D}_{t_s:t_s+\omega ,c,k},
\end{equation}
where $\cdot$ represents the dot product.

By applying Eqs.~(\ref{eq:VecWorld}) and (\ref{eq:intersection}), we can project center points of a 2D tracklet to its corresponding 3D tracklet candidates on the vertical plane, denoted as $\mathcal{T}^{2D}_{t_s:t_s+\omega ,c,k}\rightarrow \mathcal{T}^{3D}_{t_s:t_s+\omega ,c,k}$. Notably, each 3D tracklet is associated with a camera view here, and we need to further fuse multiple identical 3D tracklets into a unique one. As shown in Fig.~\ref{fig:ray2plane} (c), the 3D tracklet candidates of individuals situated on the vertical plane generally exhibit a smaller distance compared to those who are not on the vertical plane. Accordingly, we propose a plane-constraint 3D tracklet matching to select suitable 3D tracklets and fuse them. We compute the 3D Euclidean distances between cross-view projected 3D tracklet candidates, which is formulated as
	\begin{equation}
	\begin{split}  
		\mathrm{d}^{x} \left (\mathcal{T}^{3D}_{t_s:t_s+\omega ,c_i,k_i}, \mathcal{T}^{3D}_{t_s:t_s+\omega ,c_j,k_j}\right ) =~~~~~~~~~~~~~~~~~~~~~~~~~~~~~~~\\
		\left\{\begin{matrix}
		& ~\frac{\sum\limits_{\mathclap{\forall t \in \Psi_{k_i}\cap \Psi_{k_j}}}\lVert \mathbf{X}_{t,c_i,i}, - \mathbf{X}_{t,c_j,j} \rVert _{2}}{\left |\Psi_{k_i}\cap \Psi_{k_j}\right |}  , & \text{if} & \Psi_{k_i} \cap  \Psi_{k_j} \neq \emptyset ~\text{and}~ c_{i} \neq c_{j},\\ 
		& ~\emptyset, & \text{if} & \Psi_{k_i} \cap  \Psi_{k_j} =  \emptyset ,\\
		& ~\inf, &\text{if} & \Psi_{k_i} \cap  \Psi_{k_j} \neq \emptyset~\text{and}~c_{i} = c_{j} ,
		\end{matrix} \right.\\
	\forall~\mathbf{X}_{t,c_i,i} \in \mathcal{T}^{3D}_{t_s:t_s+\omega,c_i,k_i},
   \mathbf{X}_{t,c_j,j} \in \mathcal{T}^{3D}_{t_s:t_s+\omega,c_j,k_j}, ~~~~~~~~~~~~~~
	\end{split}
	\label{eq:3d_tracklets_dist}
	\end{equation}
	where $\Psi_{k_i} \in \{t_s:t_s+\omega \}$ and $\Psi_{k_j} \in \{t_s:t_s+\omega \}$ respectively denote the valid frame sets of two 3D tracklet candidates.

We apply Algorithm~\ref{algorithm} again to match 3D tracklet candidates on the vertical plane. The 3D tracklet candidates that are from different views and have an average distance bellow our clustering threshold are considered as successfully matched (see Fig.~\ref{fig:ray2plane} (d)). We fuse matched candidates, as $\big\{\mathcal{T}^{3D}_{t_s:t_s+\omega ,c_j,k_j}\mid c_j \in \mathcal{C}  \big \}\rightarrow \mathcal{T}^{3D}_{t_s:t_s+\omega,k_i}$, by averaging their 3D positions: 
\begin{equation}\label{eq:fuse3DTracklets}
	\begin{split}
	 \mathbf{X}_{t,i} =& \frac{1}{n_{cam}}\sum_{\forall c_j \in \mathcal{C} }\mathbf{X}_{t,c_j,j}, \\
	 \forall~ 
	 \mathbf{X}_{t,i} \in & \mathcal{T}^{3D}_{t_s:t_s+\omega,k_i},~
	  \mathbf{X}_{t,c_j,j} \in \mathcal{T}^{3D}_{t_s:t_s+\omega,c_j,k_j}, 
	\end{split}
\end{equation}
where $\mathcal{C}$ represents the set of matched camera views with a number of $n_{cam}$. Conversely, 3D tracklet candidates that have no cross-view pairs are treated as unmatched and subsequently discarded. Within the sliding window ranged between $t_s$ and $t_s+\omega $, selected 3D tracklets, either generated by triangulation or by ray-to-plane intersection, are combined to a set of fragmented 
3D tracklets, denoted as $\{\mathcal{T}^{3D}_{t_s:t_s+\omega ,k_0}, \mathcal{T}^{3D}_{t_s:t_s+\omega ,k_1}, \dots, \mathcal{T}^{3D}_{t_s:t_s+\omega ,k_n}\}$.

Given that our step size is half the length of our sliding window, two adjacent sliding windows share an overlap of $0.5\omega$ frames. We obtain the overlapping distances between cross-window fragmented 3D tracklets and further apply track-to-track DA~\cite{chen2020multi, chen2020cross, ohashi20vmocap, dong2021fast, rakai2022data} to connect them to complete tracklets. More specifically, we compute cross-window distance matrix $\mathbf{D}$ as
 \begin{equation}\label{eq:3d_euclidian_dist}
 \begin{split} 
	\mathbf{D}\left [k_i,k_j \right ]  &= \frac{1}{\left |\Psi_{k_i}\cap \Psi_{k_j}\right |}\sum_{\forall t \in \Psi_{k_i}\cap \Psi_{k_j}}
   \lVert \mathbf{X}_{t,i} - \mathbf{X}_{t,j} \rVert _{2},\\
   \forall~ 
   \mathbf{X}_{t,i} &\in \mathcal{T}^{3D}_{t_s:t_s+\omega ,k_i},
   \mathbf{X}_{t,j} \in \mathcal{T}^{3D}_{t_s+0.5\omega :t_s+1.5\omega ,k_j},  \\
 \end{split}
 \end{equation}
where $\mathcal{T}^{3D}_{t_s:t_s+\omega ,k_i}$ and $\mathcal{T}^{3D}_{t_s+0.5\omega :t_s+1.5\omega ,k_j}$ are cross-window fragmented 3D tracklets, $\Psi_{k_i} \in \{t_s:t_s+\omega \}$ and $\Psi_{k_j} \in \{t_s+0.5\omega :t_s+1.5\omega \}$ respectively denote their valid frame sets and $\left |\Psi_{k_i}\cap \Psi_{k_j}\right |$ is the number of valid overlapping frames.

We formulate cross-window DA as a linear assignment problem. Following conventional SC-MOT studies~\cite{DeepSORT, 1517Bochinski2017, zhang2022bytetrack, yang2023hard, luiten2020track, kim2021eagermot}, we use a Boolean matrix $\mathbf{M}$ to model the cross-window track-to-track assignments. When row $i$ is assigned to column $j$, the matrix element $\mathbf{M}_{i,j}$ is set to 1. Each row is assigned to at most one column, and each column is assigned to at most one row. Consequently, the optimal assignment $\mathbf{M}^*$ is obtained by minimizing the total cost, as described below:
 \begin{equation}\label{eq:hungarian}
   \begin{split} 
 \mathbf{M}^* =& \argmin_{\mathbf{M}} \sum_{i,j} \mathbf{D}_{i,j}  \mathbf{M}_{i,j}\\
 \textrm{s.t.} \sum_{i} \mathbf{M}_{i} =& 1~\textrm{and}~\sum_{j} \mathbf{M}_{j} = 1 . 
   \end{split}
 \end{equation}

Based the optimized $\mathbf{M}^*$, the matched cross-window 3D tracklets are connected. For instance, when cross-window 3D tracklets $\mathcal{T}^{3D}_{t_s:t_s+\omega ,k_i}$ and $\mathcal{T}^{3D}_{t_s+0.5\omega :t_s+1.5\omega ,k_j}$ are associated, their fused tracklet $\mathcal{T}^{3D}_{t_s:t_s+1.5\omega ,k_i}$ can be updated as
	\begin{equation}\label{eq:temporalConnection}
	  \begin{split} 
		\mathbf{X}_{t,k_i} =& 
		\begin{cases}
			\mathbf{X}_{t,k_i} , & \text{for } t_s \leq t < t_s+0.5\omega  \\
		\left (\mathbf{X}_{t,k_i} + \mathbf{X}_{t,k_j} \right )/2, & \text{for } t_s+0.5\omega  \leq t < t_s+\omega  \\
		\mathbf{X}_{t,k_j} , & \text{for } t_s+\omega  \leq  t  <  t_s+1.5\omega  \\
		\end{cases} \\
		\forall ~\mathbf{X}_{t,k_i}  \in & \mathcal{T}^{3D}_{t_s:t_s+1.5\omega ,k_i}, ~~\mathbf{X}_{t,k_j} \in \mathcal{T}^{3D}_{t_s+0.5\omega :t_s+1.5\omega ,k_j}.
	  \end{split}
	\end{equation}

In contrast, the unmatched 3D tracklets from the new sliding window are considered as new tracklets and are assigned with new track IDs. Once all sliding windows have been processed, fragmented 3D tracklets are associated to complete 3D tracklets. Meanwhile, we propagate the correlation of 3D tracklets to multi-view 2D tracklets and connect them in the same approach. This ensures a consistent and coherent representation of both 3D and 2D tracklets throughout the cross-window DA.

\subsection{Identify Target Gymnast for Pose Estimation}

Safety is a top priority in gymnastics, and assistants are allowed to enter the performing area to protect the gymnast in case of an emergency. Therefore, multiple 3D tracklets may be present within  our pre-defined performance space $\boldsymbol{S}^{perf}$. As our analysis indicates, it is challenging to identify the target gymnast by textures, because the roles of the target athlete and the assistant may be interchangeable during daily training, and new uniforms may be introduced at gymnastics events. To address this problem, we identify the target gymnast by referring to their 3D tracklets.

We utilize the central positions of tracklets in the aforementioned cross-window cross-view matching. However, upon completion of the matching, we also compute 3D top/bottom positions of tracklets, $\mathbf{X}_{t,k}^{top}$ and $\mathbf{X}_{t,k}^{bot}$, using multi-view 2D top/bottom positions of associated cross-view 2D bboxes. They will then be used to identify target gymnasts. In gymnastics performances, it is assumed that the 3D top/bottom positions of the target gymnast can rapidly reach certain heights once the performance begins~\cite{de20172020, de2018technical}. Within the performance space $\boldsymbol{S}^{perf}$, we therefore establish height thresholds for 3D top/bottom positions, denoted as $h^{top}$ and $h^{bot}$, to trigger the identification of the target gymnast. The function to identify the target athlete tracklet is formulated as follows:
\begin{equation}\label{eq:identify}
	\resizebox{\linewidth}{!}{$
	\begin{split}
		\textrm{identify}\left (\mathcal{T}^{3D}_{t_c-\Delta :t_c, k_i}, \boldsymbol{S}^{perf}, h^{top}, h^{bot} \right ) =~~~~~~~~~~~~~~~~~~~~~~~~~~~~~~~~~~~~~~~~~~~~~~~~~~~~~~~~~~~~\\
		\begin{cases}
			1, & \text{if}~ \sum\limits_{t \in  [t_c-\Delta:t_c] } \mathbb{1} \left [ \left ( \mathbf{X}_{t,k_i} \in \boldsymbol{S}^{perf} \right ) \wedge \left ( \mathbf{X}_{t,k_i}^{top}[2]>h^{top} \right ) \wedge \left ( \mathbf{X}_{t,k_i}^{bot}[2]>h^{bot} \right ) \right ] > 0.5\Delta,  \\
			0, & \text{otherwise},
		\end{cases} \\
		\forall  ~\mathbf{X}_{t,k_i}, \mathbf{X}_{t,k_i}^{top}, \mathbf{X}_{t,k_i}^{bot} \in  \mathcal{T}^{3D}_{t_c-\Delta:t_c, k_i}, ~~~~~~~~~~~~~~~~~~~~~~~~~~~~~~~~~~~~~~~~~~~~~~~~~~~~~~~~~~~~~~~~~
	\end{split}
	$}
\end{equation}
where $\mathbb{1}$ denotes an indicator function to verify if a 3D position satisfies certain conditions, $\Delta$ is a temporal buffer and we experimentally set $\Delta=30$.

When the target gymnast has not been assigned, but 3D positions exist in the performance space $\boldsymbol{S}^{perf}$, we apply Eq.~\ref{eq:identify} to identify the tracklet ID associated with the target gymnast.
We do not apply Eq.~\ref{eq:identify} for every frame. Once the target gymnast has been initialized, the same track ID generated by MOT is used to identify the target gymnast until an ID switch occurs. Note that, although we start to track a single target gymnast after the initial identification, we continue to apply MOT instead of single-object tracking (SOT). This choice is motivated by the fact that, when non-target detections are close to the target tracklet, using SOT may result in a sub-optimal matching and lead to ID switches. In contrast, employing MOT allows for a joint optimization for all individuals, thereby yielding globally optimized tracking results. 

Although we have introduced gymnastics domain knowledge to improve the robustness of gymnast tracking, ID switches still occur. This typically happens when
\begin{itemize}
	\item Only one-camera detections are available for the target gymnast and the corresponding 3D positions cannot be generated. 
	\item In extremely crowded scenes, a large number of 2D tracklets may generate pseudo 3D tracklet candidates that are randomly located in the performance space, which causes an incorrect data association.
\end{itemize}
Whenever an ID switch happens, our tracking system loses the target within the current sliding window, and thus, it will attempt to re-identify the target gymnast from 3D tracklet candidates in the subsequent sliding window with Eq.~\ref{eq:identify}. The re-identified target 3D tracklet must meet our pre-defined criteria: it should reach the trigger heights and remain within the performing space for multiple shots. These criteria are the same as those we used to identify the target gymnast at the beginning of the performance.  Although this process may immediately halt incorrect tracking, we inevitably lose some frames in the gymnast tracking. To compensate for these missing frames, we will perform spatiotemporal interpolation to fill in the missing gap if it is within $7$ frames.

After obtaining the target gymnast, we apply interpolation and smoothing to the target gymnast's 3D tracklet. The refined 3D tracklet is then projected onto each 2D view, optimizing the 2D tracklets with multi-view information. We utilize the multi-view 2D tracklets of the target gymnast to estimate 3D poses. Pre-identifying the target gymnast greatly eliminates computational cost since non-target individuals have been excluded, and the high-resolution image can be used to improve the pose estimation performance. 
In pose estimation, there are crucial aspects that need to be considered. For instance, due to potential detection bias, the 2D bbox may not fully encompass the target gymnast, which can lead to pose estimation errors. To mitigate these errors, we expand each bbox of 2D tracklets to accommodate for detection bias, as
	\begin{equation}\label{eq:bboxResize}
	  \begin{split} 
		\boldsymbol{b}_{t,c,i} =& \left [x_{t,c,i}, y_{t,c,i}, w_{t,c,i}, h_{t,c,i} \right ] \\
	\Rightarrow \boldsymbol{b}^{buf}_{t,c,i} =& \left [x_{t,c,i}, y_{t,c,i}, \alpha \ell, \alpha \ell \right ], ~\textrm{with}~\ell = \textrm{max}\left (w_{t,c,i}, h_{t,c,i} \right )
	  \end{split}
	\end{equation}
where $\alpha$ is the expanded scale and we experimentally set $\alpha=1.3$. In our setting, $[x_{t,c,i}, y_{t,c,i}]$ are the center point other than the top-left point of the bbox, which simplifies computations in our cross-view matching and 2D pose estimation.

\begin{table*}[h!]
	\footnotesize	
	\centering
	\caption{\textbf{Comparison of Multi-camera Tracking Dataset.} Our gymnast tracking distinguishes itself from others since the gymnast may move to arbitrary heights and perform various poses. \revised{These factors significantly increase the challenges associated with cross-camera data association and 3D tracklet generation.}}
	\label{table:datasets}
	\setlength{\tabcolsep}{.1mm}{
	\begin{tabular}{lccccccc}
		\textbf{Dataset}   & \textbf{Video Resolution} & \textbf{Video FPS} & \textbf{Video Length}  & \textbf{No. Cameras}~~ & \textbf{Tacking Target Bottom Position} & \textbf{Pose Diversity} \\ \toprule
\begin{tabular}[c]{@{}l@{}}EPFL(Laboratory,\\ Passageway, \revised{Basketball})~\cite{fleuret2007multicamera}\end{tabular} 
& $320\times240$ & 25 & 26 min.  & 4 & On the Ground Plane (0 m)  &  Run, Walk, Stand \\
PETS 2009~\cite{ferryman2009pets2009} & $768 (720)\times576$ & 7  & 2 min.  & 8  & On the Ground Plane (0 m) &  Walk, Stand \\
Campus~\cite{xu2016multi} & $852 \times 480$  & 25  & 4 $\times$ 4   min.  & 4   & On the Ground Plane (0 m) &  Walk, Stand \\
	WildTrack~\cite{chavdarova2018wildtrack} & $1920\times1080$ & 60  & 60 min.  & 7 & On the Ground Plane (0 m)  &  Walk, Stand \\
	\revised{MMPTRACK}~\cite{han2023mmptrack} & $1920\times1080$ & 15  & 576 min.  & 23 & On the Ground Plane (0 m)  &  Walk, Stand \\
	\textbf{Ours}  & $1920\times1080$  & 30  & 140 $\times$ (1$\sim$8)  min. & 4   & Arbitrary Heights (0$\sim$3 m) &  
	\begin{tabular}[c]{@{}c@{}}  Jump, Twist, Squats, \\ Push-ups, \etc \end{tabular}\\
	\bottomrule
	\end{tabular}
	}
\end{table*}

Utilizing the expanded bounding boxes, we incorporate a fine-tuned HRNet~\cite{sun2019deep} to generate 2D poses for each camera view, and then apply Eq.~(\ref{eq:3d_triangulation}) to estimate the 3D poses in a computationally efficient manner. Given sufficient computing resources, a volumetric triangulation method~\cite{iskakov2019learnable} can be employed in conjunction with HRNet to achieve a more accurate 3D pose estimation. In the final stage, the 3D pose sequence of the target gymnast is segmented and used to generate frame-wise gymnastics code of points
for judging purposes.

\subsection{ Configurations for our tracking framework.} 
\noindent{\bf Input sizes of video frames.} Our original videos have a resolution of $1080\times1920$. To accelerate the processing speed, each frame is rescaled and padded to $608\times608$ resolution for efficient person detection. After acquiring the multi-view 2D tracklets of the target gymnast, we buffer original 2D bbox $\boldsymbol{b}_{t,c,i}$ to $\boldsymbol{b}^{buf}_{t,c,i}$ and use it to crop a patch on the $1080\times1920$ image. The high-resolution patch enables better pose estimation for the target gymnast.

\noindent{\bf IoU tracker parameters.} In our single-view tracking, we apply IoU trackers~\cite{1517Bochinski2017} to associate 2D bboxes to 2D tracklets. In all our gymnastic routines, we set the IoU threshold to be $0.1$, which is sufficient to handle the irregular movements of gymnasts at a video recording speed of $30$ FPS. Moreover, informed by the analysis from the previous work, ReMOT~\cite{yang2021remot}, we prioritize cut-off errors over mix-up errors in our 2D tracklets. This preference stems from our capability to correct cut-off errors in our cross-window cross-view DA. Consequently, we set the maximum tracking age to $2$, a decision that may increase cut-off errors but simultaneously reduce mix-up errors.

\begin{figure}[h!]
	\centering	\includegraphics[width=\columnwidth]{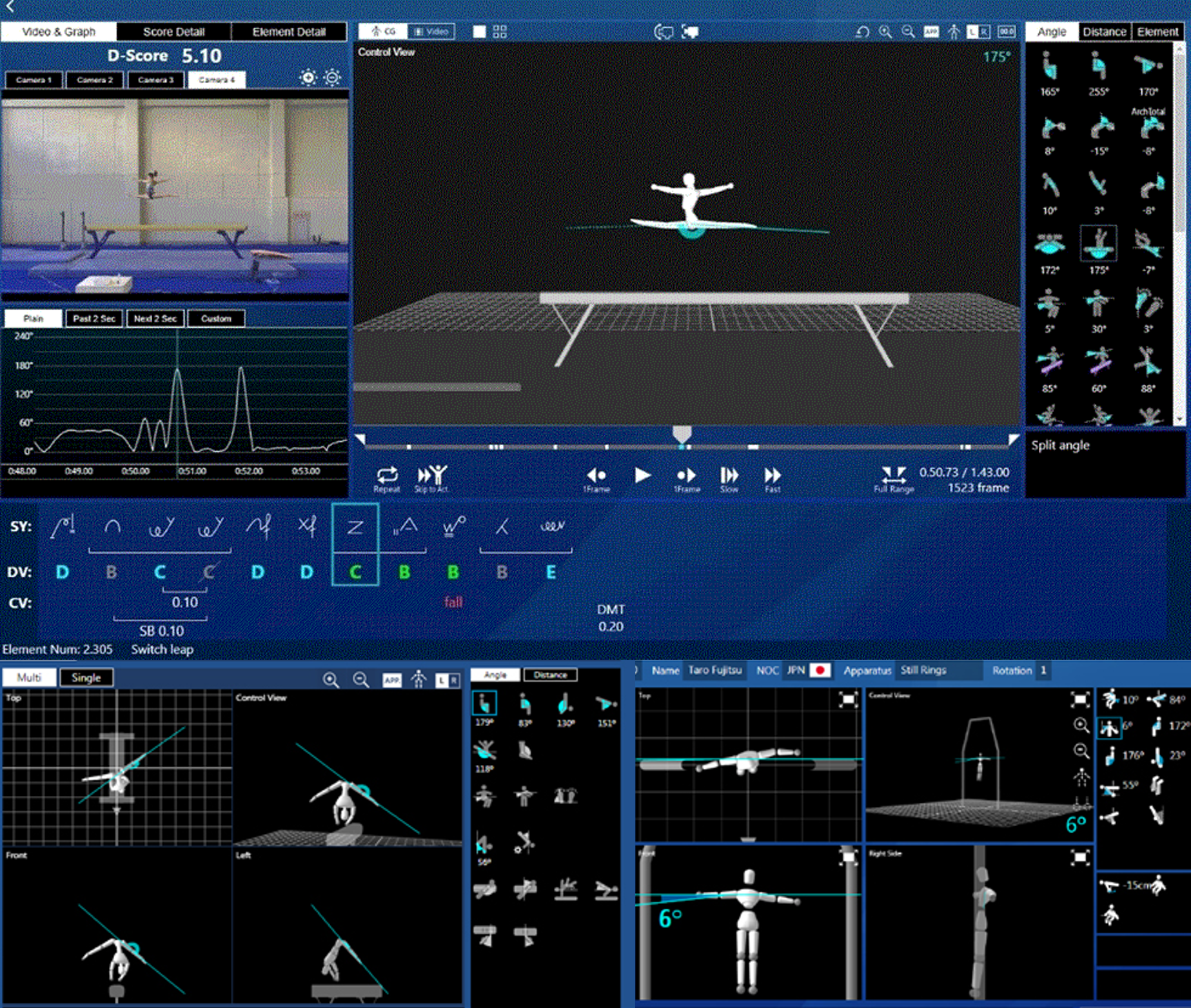}%
	\caption{ \textbf{The display of our gymnastics judging support system.}  Referring to angles within poses, a segment of the 3D pose sequence is assigned to the gymnastics code of points~\cite{de20172020, de2018technical, palmer2022aesthetics} to which it corresponds.}
	\label{fig:score}
\end{figure}

\noindent{\bf Cross-window cross-view DA parameters.} We adhere to the best practice of UniMMT~\cite{yang2023unified}, establishing the threshold $\lambda$ for associating multi-view 2D tracklets at $0.3$. For our novel module designed to match coplanar 3D tracklet candidates, we experimentally set the threshold at $0.5$ meters. Additionally, for our cross-window DA, the unmatched threshold is determined to be $0.6$ meters.

\subsection{Gymnastics Judging Module}

Although the gymnastics judging module is not the focus of this paper, we provide a simple introduction to it for promoting understanding of the gymnastics judging system for which this tracking framework is used. Other than giving a final score for the gymnastic performance, gymnastic judging is more akin to video event segmentation and recognition~\cite{yang2023weakly}. In general, a complete gymnastic performance comprises multiple temporal pieces of skills. Each skill, depending on its category and quality, can be assigned a temporal label corresponding to the gymnastics code of points~\footnote{\url{https://www.youtube.com/watch?v=9kDD0DlDiA4}}~\cite{de20172020, de2018technical, palmer2022aesthetics}. The manual judging progression can be learned from a third-party video available on YouTube~\footnote{\url{https://www.youtube.com/watch?app=desktop&v=Hnr2txGcRL0}}. Our gymnastic judging system has automated this process~\cite{ikeda2022skeleton, asayama2023skeleton}, as depicted in Fig.~\ref{fig:score}. Coaches and trainers can utilize the outputs (\ie, gymnastics code of points) to analyze and evaluate the performance with greater precision, providing valuable feedback for improvement.

\section{Experiments}

\subsection{Experimental Settings}

\noindent{\bf Evaluation dataset.} 
To evaluate our proposed gymnast tracking methodology, we have created an extensive gymnastics dataset that encompasses a diverse range of gymnastics types from daily training. This dataset includes the Balance Beam (BB), Pommel Horse (PH), Vault (VT), Still Rings (SR), Horizontal Bar (HB), Uneven Bars (UB), and Parallel Bars (PB). We have gathered data from $20$ unique performances for each category of gymnastics, culminating in a total of $140$ performances that exhibit a broad and diverse range. To obtain the ground truth for our evaluation, we have carefully reviewed and corrected biases present in the tracklets. 

We conducted a comparison of our dataset with previously existing ones, as illustrated in TABLE~\ref{table:datasets}. Most of the existing multi-camera tracking datasets predominantly cover only two types of activities---``Walk'' and "Stand"---and they typically feature the person's bottom point on the ground plane. In contrast, our dataset includes a gymnast performing challenging activities such as ``Jump'', ``Twist'', ``Squats'', etc., with the bottom point at varying heights. When the bottom points of tracking targets are situated on the ground at zero height, the corresponding 3D footprints can be readily generated by referring to the bottom point of a monocular view through homograph projection~\cite{fleuret2007multicamera, yoo2016online, chen2016equalized, wen2017multi, chen2020multi,he2020multi, nguyen2022lmgp, luna2022graph, huang2023enhancing}. The cross-camera matching can then be easily approached by clustering 3D footprints that are mapped from different cameras. However, when the bottom points of tracking targets are potentially at arbitrary heights, as is the case with our gymnast tracking, the direct computation of 3D footprints becomes unfeasible. To solve this problem, we need to integrate more multi-camera spatiotemporal geometric relationships, as introduced in our framework.

\begin{figure}[h!]
	\centering	\includegraphics[width=0.9\columnwidth]{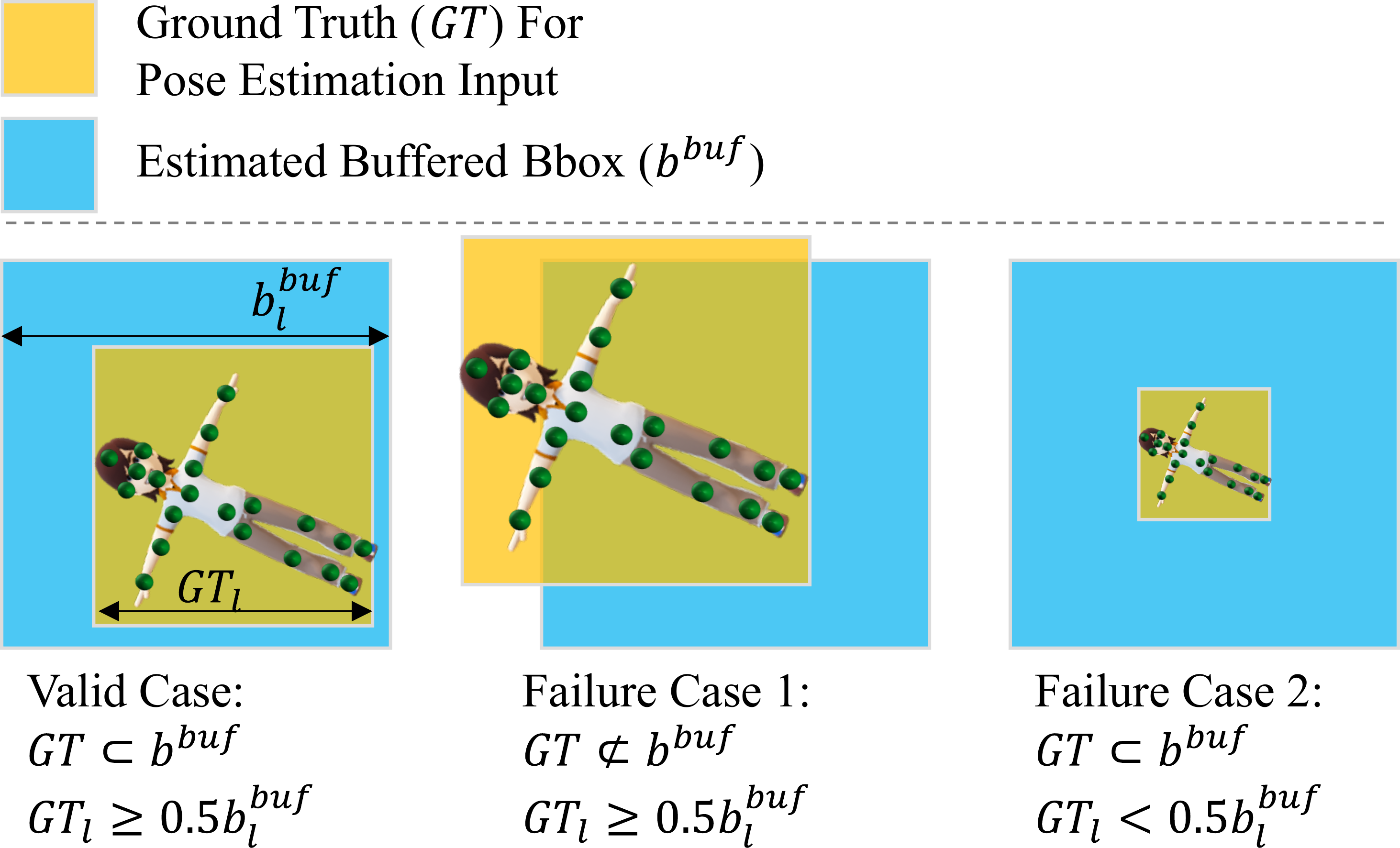}
	\caption{ \textbf{Evaluating the suitability of 2D bbox for pose estimation.} Whenever the ground-truth bbox length is less than half the $\boldsymbol{b}^{buf}$ or lies outside it, the $\boldsymbol{b}^{buf}$ is considered a failure because it impairs the pose estimation performance. }
	\label{fig:metrics}
\end{figure}

\noindent{\bf Evaluation metrics.} 
Owing to the unique nature of our gymnast tracking, we apply MOT during the process, but our final results are the 3D tracklets and corresponding multi-view 2D tracklets of the target gymnast. Hence, conventional MOT metrics, like HOTA~\cite{HOTA} and MOTA~\cite{bernardin2008evaluating}, are unsuitable for our tracking evaluation. We opt to use three metrics to evaluate the 3D tracklet and corresponding multi-view 2D tracklets for the target gymnast. Our first metric is the number of ID switches, which has been applied in the conventional MOT evaluation~\cite{DeepSORT, 1517Bochinski2017, zhang2022bytetrack, yang2023hard,  luiten2020track, kim2021eagermot} but is also appropriate in our case. Our second metric compares the average Euclidean distance (AED) between the estimated 3D central positions of target 3D tracklet to the ground truth. Our last metric measures the failure rate (FR) of target multi-view 2D tracklets by assessing their suitability for 2D pose estimation. As Fig.~\ref{fig:metrics} illustrates, whenever the ground-truth bbox length is less than half the $\boldsymbol{b}^{buf}$ or lies outside it, the $\boldsymbol{b}^{buf}$ is considered a failure because it impairs the pose estimation performance. While these metrics provide insights into tracking performance from various perspectives, they are also correlated with each other.

\noindent{\bf Baseline trackers.} 
We adopt state-of-the-art multi-camera MOT algorithm MvPT~\cite{dong2021fast} and UniMMT~\cite{yang2023unified} as our baseline tracking algorithm. While MvPose~\cite{dong2019fast} is a pioneering work that contributes cross-view multi-person DA at arbitrary heights, MvPT builds upon it to enable its tracking capability. To handle weak geometric cues, UniMMT further extends MvPT by jointly associating cross-view 2D tracklets. The effectiveness of UniMMT has been demonstrated through comparisons with state-of-the-art methods~\cite{zhang2022voxeltrack, han2023mmptrack} across various public datasets. Our proposal partially inherits from UniMMT. The primary distinction between our proposal and baseline methods lies in our incorporation of gymnastics domain knowledge with a novel cascaded DA, which aims to enhance the robustness of our gymnast tracking. Although incorporating the appearance cue could enhance tracking performance, we employ only the 2D bbox center to form geometric cues in our evaluation to prioritize computational efficiency for MvPT, UniMMT, and our proposal.

\begin{table*}[h!]
	\scriptsize
	\centering
	\caption{\textbf{Comparisons of tracking performance using various camera settings on our multi-camera gymnast tracking dataset.} We define the number of views with valid detections as $n_{cam}$. We specifically select opposite-view cameras for $n_{cam}=2$. The best result is rendered with \textbf{bold values}. Results with significant bias are highlighted in \textcolor{red}{red}.}
	\label{table:compareTracking}
	\resizebox{\linewidth}{!}{
	\setlength{\tabcolsep}{.01mm}{
	\begin{tabular}{lrrrrrrrrrrrrrrrrrrrrr}   
	\toprule
	\textbf{Method} & \multicolumn{3}{c}{\textbf{Balance Beam (BB)}
	} & \multicolumn{3}{c}{\textbf{Pommel Horse (PH)}
	} & \multicolumn{3}{c}{\textbf{Vault (VT)}} & \multicolumn{3}{c}{\textbf{Still Rings (SR)}
	} & \multicolumn{3}{c}{\textbf{Horizontal Bar (HB)}
	} & \multicolumn{3}{c}{\textbf{Uneven Bars (UB)}
	} & \multicolumn{3}{c}{\textbf{Parallel Bars (PB)}
	} \\ 
	\cmidrule(lr){2-4} \cmidrule(lr){5-7} \cmidrule(lr){8-10} \cmidrule(lr){11-13} \cmidrule(lr){14-16} \cmidrule(lr){17-19} \cmidrule(lr){20-22}
	& $n_{cam}$=2,  & ~~~3,&  4~ & $n_{cam}$=2,  & ~~~3,&  4~ & $n_{cam}$=2,  & ~~~3,&  4~ & $n_{cam}$=2,  & ~~~3,&  4~ & $n_{cam}$=2,  & ~~~3,&  4~  & $n_{cam}$=2,  & ~~~3,&  4~ & $n_{cam}$=2,  & ~~~3,&  4~\\
	\midrule
	\multicolumn{22}{l}{\textbf{Evaluate the number of ID Switches for target 3D (2D) tracklets (the smaller represents the better):}}\\
	MvPT (w/o Re-ID)\cite{dong2021fast} & 12  & 6  & 5 & 5  & 1 & 1  & 6 & 3  & 1  &\textbf{2}&\textbf{0}&\textbf{0} &3& 3 &1& 19  & 8   & 8 & 8   &3  &2\\
	UniMMT (w/o Re-ID)\cite{yang2023unified} & 7  & 2 & 2& 2  &\textbf{0}&\textbf{0}& 1   & 1 &1&\textbf{2}& \textbf{0} &\textbf{0} & 2 &\textbf{0}&\textbf{0}& 13  & 5   & 3 & 6  &1 &1\\
	\textbf{Ours} \revised{(w/o Re-ID)}  & \textbf{2} &\textbf{1}&\textbf{1}& \textbf{1} &\textbf{0}&\textbf{0}& \textbf{0} &\textbf{0}&\textbf{0}& \textbf{2} &\textbf{0}&\textbf{0} & \textbf{1} &\textbf{0}&\textbf{0}& \textbf{2} & \textbf{2}   & \textbf{2} & \textbf{3} &\textbf{0}&\textbf{0}\\\hline\\
	\multicolumn{22}{l}{\textbf{Evaluate central positions of 3D tracklets with AED in meters (the smaller represents the better):}}\\
	MvPT (w/o Re-ID)\cite{dong2021fast} & 0.258  & ~ 0.066  & ~ 0.058 & 0.317   & 0.099  & 0.092  & 0.261   & 0.065  & 0.060  & 0.098  & 0.087 & 0.081    & 0.217  & 0.103  & 0.090   & \textcolor{red}{2.741}  & 0.179   & 0.163 & 0.205   &0.103 & 0.101\\
	UniMMT (w/o Re-ID)\cite{yang2023unified} & 0.242  & 0.061  & 0.053 & 0.301  & 0.093  & 0.088 & 0.243   & 0.051  & \textbf{0.044}  & \textbf{0.084}  & \textbf{0.078}  & \textbf{0.076}   & 0.184  & 0.093 & \textbf{0.082}   & \textcolor{red}{2.711}  & 0.115   & 0.096 & 0.176   &0.089 & 0.080\\
	\textbf{Ours} \revised{(w/o Re-ID)}   & \textbf{0.160} & \textbf{0.054}  & \textbf{0.050}  & \textbf{0.236}  & \textbf{\textbf{0.091}}  & \textbf{0.086}  & \textbf{0.174}  & \textbf{0.049}   & \textbf{0.044}    & 0.086 & \textbf{0.078}  & \textbf{0.076}   & \textbf{0.168}  & \textbf{0.087}  & \textbf{0.082}  & \textbf{0.232}  & \textbf{0.091}   & \textbf{0.089} & ~\textbf{0.095}  &~\textbf{0.084} &~\textbf{0.078} \\\hline\\
	\multicolumn{22}{l}{\textbf{\textbf{Evaluate 2D bboxes of valid multi-view 2D tracklets with Failure Rate (the smaller represents the better):}}}\\
	MvPT (w/o Re-ID)\cite{dong2021fast} & \textcolor{red}{3.14\%}  &0.18\% & 0.15\% & 0.64\%   & 0.16\%  & 0.14\%   & 0.30\%    & 0.15\% & 0.12\%  & \textbf{0.05}\%  & \textbf{0.04}\%   & \textbf{0.04}\%    & 0.12\%  & 0.10\%  & 0.08\%  & \textcolor{red}{4.01\%}  & 0.36\%  & 0.25\%  & 0.34\%   &0.12\% &0.08\%\\
	UniMMT (w/o Re-ID)\cite{yang2023unified} & \textcolor{red}{2.20\%}  &0.12\%  & 0.09\%  & 0.39\%   & 0.13\%  & \textbf{0.11}\%   & 0.28\%  & 0.08\%  & 0.08\%  & \textbf{0.05}\%   & \textbf{0.04}\%  & \textbf{0.04}\%   & \textbf{0.07}\% & \textbf{0.06}\% & \textbf{0.06}\%   & \textcolor{red}{3.85\%}  & 0.18\%    & 0.15\%  & 0.32\%   &0.10\% &0.07\%\\
	\textbf{Ours} \revised{(w/o Re-ID)} & \textbf{0.27}\%  & \textbf{0.10}\%  & \textbf{0.08}\%  & \textbf{0.22}\%   & \textbf{0.11}\%  & \textbf{0.11}\%  & \textbf{0.16}\%   & \textbf{0.07}\%   & \textbf{0.07}\% & \textbf{0.05}\%   & \textbf{0.04}\%   & \textbf{0.04}\%   & \textbf{0.07}\%   & \textbf{0.06}\% & \textbf{0.06}\%  & \textbf{0.38}\%  & \textbf{0.12}\%    & \textbf{0.10}\% & \textbf{0.18}\%   &\textbf{0.05}\% &\textbf{0.05}\%\\
	\bottomrule
\end{tabular}}
	}
\end{table*}

\begin{table*}[h!]
	\footnotesize	
	\centering
	\caption{\revised{\textbf{Comparisons of average tracking failure rate and speed on our dataset.} The best and the second-best results are rendered with \textbf{bold values} and \underline{underline}, respectively. The detector running time is not included. Appearance feature (\ie, w/ Re-ID) can be used to improve the tracking results but is time-consuming. }}
	\label{table:speed}
	\begin{tabular}{lcccc}
	\toprule  
	\textbf{Method} & \multicolumn{3}{c}{\textbf{Avg. Failure Rate}}& \textbf{Avg. Tracking Speed (FPS)} \\ 
	\cmidrule(lr){2-4} 
	& $n_{cam}$=2,  & ~~~3,&  4~ &
	\\\hline
	MvPT (w/ Re-ID)\cite{dong2021fast}	& 1.26\%  & 0.14\%  & 0.10\% &  7          \\
	UniMMT (w/ Re-ID)\cite{yang2023unified}	  &1.08\% &  0.10\% & \underline{0.08\%}   & 21         \\ 
	\textbf{Ours} (w/ Re-ID)  & \textbf{0.18}\% &  \textbf{0.07}\% & \textbf{0.07}\% &    18      \\ \hline
	MvPT (w/o Re-ID)\cite{dong2021fast}	& 1.34\%  & 0.16\%  & 0.12\%  &  223          \\
	UniMMT (w/o Re-ID)\cite{yang2023unified}	 &1.11\% &  0.12\% & 0.09\%   & \textbf{468}         \\ 
	\textbf{Ours} (w/o Re-ID) & \underline{0.19\%} &  \underline{0.08\%} & \textbf{0.07}\% &   \underline{425}      \\
	\bottomrule    
\end{tabular}
\end{table*}

\subsection{Experimental Results}

\noindent{\bf Overall performance comparisons for gymnast tracking.} 
As shown in TABLE~\ref{table:compareTracking}, we quantitatively compare our gymnast tracking results with previous works, MvPT~\cite{dong2021fast} and UniMMT~\cite{yang2023unified}, under various conditions: two opposing camera views (\ie, $n_{cam}=2$), three camera views (\ie, $n_{cam}=3$), and a full setting with four camera views (\ie, $n_{cam}=4$). Note that, due to potential detection failures, having $n_{cam}=4$ does not necessarily imply that the number of available cross-camera detections is $4$. Instead, it could fall within the range of $\{0, 1, 2, 3, 4\}$. This same principle applies to instances where $n_{cam}=2$ and $n_{cam}=3$.

Overall, from the results of TABLE~\ref{table:compareTracking}, we have demonstrated the robustness of our method by introducing the domain knowledge of gymnastics. In detail, when compared to UniMMT, our method produces results that align closely when $n_{cam}=4$, are similar when $n_{cam}=3$, and significantly diverge when $n_{cam}=2$. Because our tracking method is partially inherited from UniMMT, our cascaded DA reverts to UniMMT's DA when detections from at least three views are valid. With $n_{cam}=4$, we assume that our overall detection failures constitute only a minor fraction, and thus, our cascaded DA reverts to UniMMT's DA for most of frames, resulting in closely aligned outcomes. However, when $n_{cam}=2$, our cascaded DA demonstrates greater robustness than UniMMT's DA, enabling us to achieve comparable or superior performance in the seven types of gymnastics listed. Unlike our method and UniMMT, MvPT typically employs single-frame full-body keypoints for cross-view DA. However, for computational efficiency, we require all methods to rely solely on a weak geometric cue: the central points of bounding boxes. This leads to an increased number of ID switches and a larger bias in both 2D and 3D tracklets for MvPT. In contrast, our method and UniMMT utilize segmented 2D tracklets to provide more robust geometric cues for cross-view DA. Consequently, our method and UniMMT outperform MvPT in most of the experimental results.

\begin{figure*}[h!]
	\begin{center}
	\hsize=\textwidth
	\includegraphics[width=\textwidth]{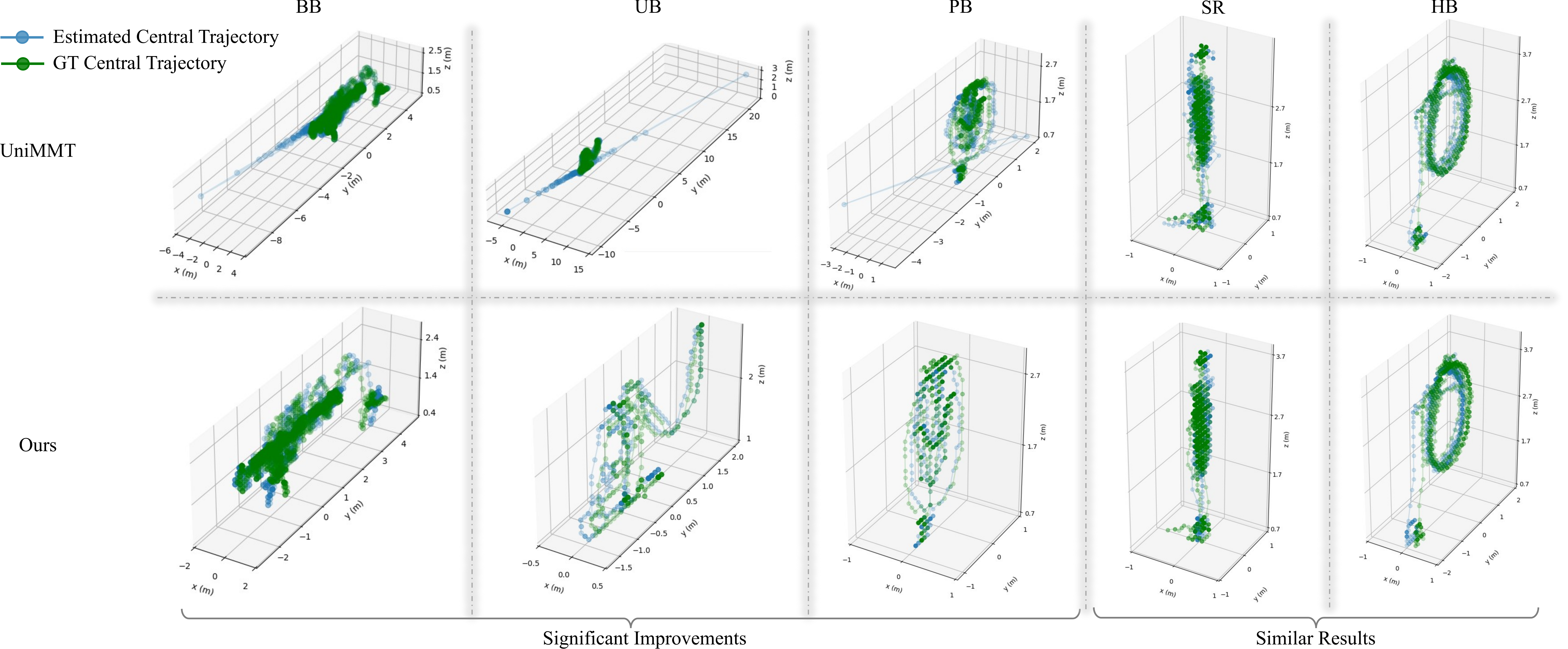}
			\caption{\textbf{Comparison of 3D tracklet centers between UniMMT~\cite{yang2023unified} and our method.} Valid detections are only available in two opposing views (\ie, $n_{cam}=2$).}
	\label{fig:comparation_plots}
\end{center}
\end{figure*}

\begin{figure*}[h!]
	\begin{center}
		\hsize=\textwidth
		\includegraphics[width=\textwidth]{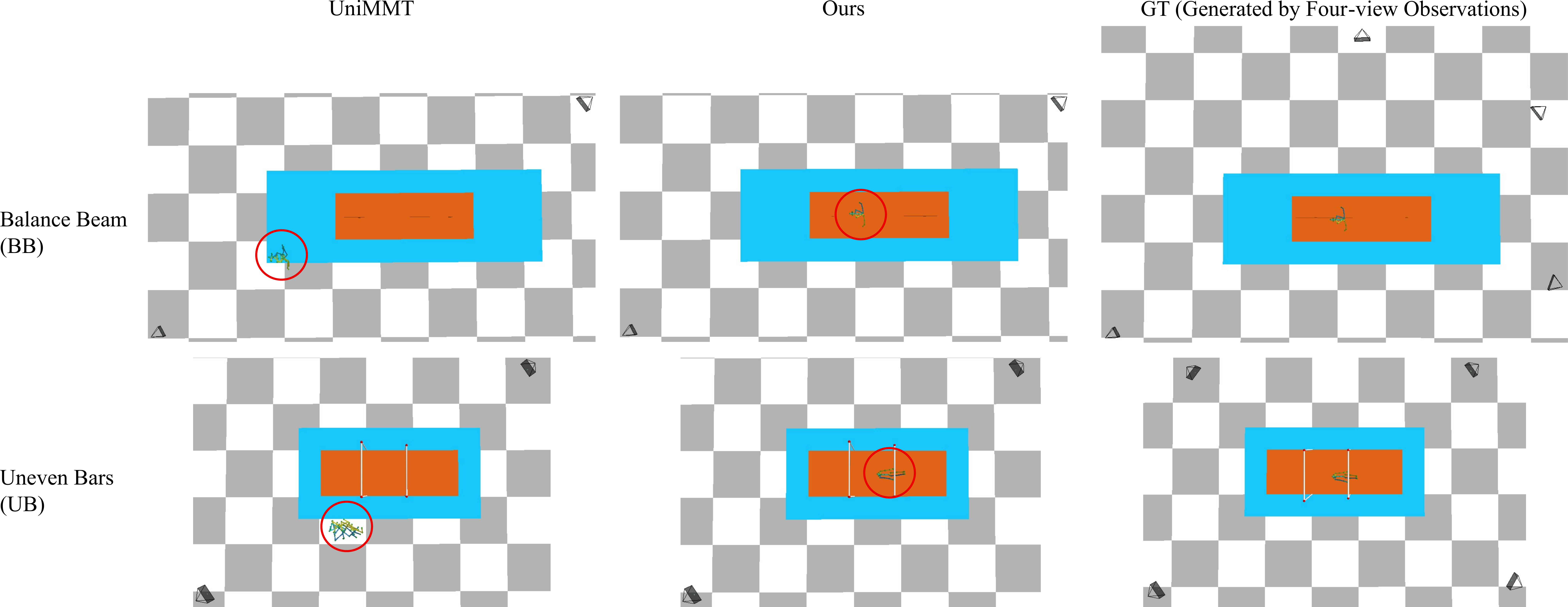}
				\caption{\revised{\textbf{Comparison of estimated 3D gymnast poses using UniMMT~\cite{yang2023unified} tracking outputs versus ours.} For the difficult cases of BB and UB, we intend to block the observations of two cameras and use the corresponding tracking results for pose estimation. The final results are compared with the GT generated by using four cameras. }}
	\label{fig:pose_compare}
	\end{center}
	\end{figure*}

\begin{figure*}[h!]
	\begin{center}
		\hsize=\textwidth
		\includegraphics[width=\textwidth]{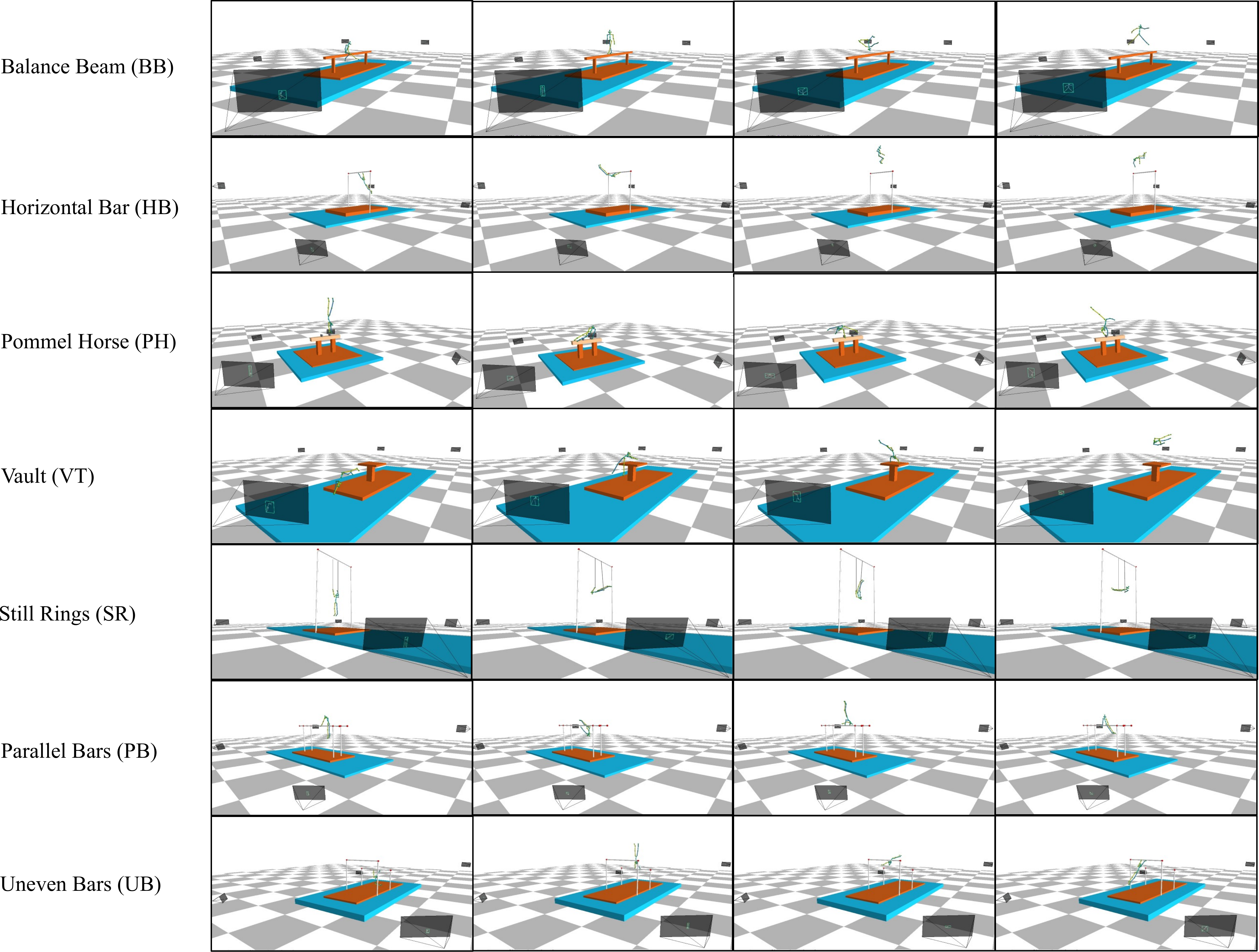}
				\caption{\textbf{Estimated 3D poses for our tracked target gymnast.} High-quality tracklets are obtained by using our gymnast tracking and accurate 3D poses can be estimated.}
	\label{fig:demo}
	\end{center}
	\end{figure*}

Compared with MvPT and UniMMT, our method specifically incorporates gymnastics domain knowledge to enhance the robustness in scenarios where detections from two opposing views are insufficient. Notably, in challenging UB routines with $n_{cam}=2$, our method reduces ID switches by $89.5\%$ and $84.6\%$, ADE by $91.5\%$ and $91.4\%$, and FR by $90.5\%$ and $90.1\%$ respectively, when compared to MvPT and UniMMT. The only exception is the SR routines, where all methods exhibit similar performances when $n_{cam}=\{2, 3, 4\}$. This can be attributed to the gymnast's solo performance at a high position with few occlusions, and two facing-upwards cameras have a relatively smaller line-of-sight angle. Although our AED value is slightly higher than UniMMT's in SR routines when $n_{cam}=2$, it falls within an acceptable range and does not significantly affect the downstream pose estimation. Furthermore, in real practice, we generally have valid detections over three views in most frames, and thus our method could perform as well as UniTMM in SR, while guaranteeing better performance in other challenging routines.

\noindent{\bf Trade-off between tracking performance and speed.} \revised{Although it is a common practice to integrate both geometric and appearance features to achieve high-performance MOT solutions~\cite{wen2017multi, he2020multi, you2020multi, nguyen2022lmgp, huang2023enhancing, du2023StrongSORT}, extracting appearance features using re-identification (Re-ID) models~\cite{li2019attribute, qi2020progressive, tao2023adaptive, he2023fastreid} incurs additional computational costs. Given that our gymnastics judging system demands real-time processing and involves multiple tasks---including detection, gymnast tracking, pose estimation, and action segmentation---a highly efficient tracking solution is imperative.}

\revised{To rigorously assess the trade-off between tracking performance and speed, we conduct a comparative analysis of the average tracking failure rate and speed on our dataset, with and without using appearance features. The experimental results are shown in TABLE~\ref{table:speed}. All methods use only the central points of bounding boxes as keypoints for cross-camera and cross-frame DA. We follow previous MOT works (\eg,~\cite{zhang2022bytetrack}) to  use FastReID~\cite{he2023fastreid} to extract appearance features. 
The results indicate that utilizing appearance features (w/ Re-ID) can decrease the average tracking failure. However, while this decrease in failure rate is marginal, the reduction in speed is considerable, nearly reducing performance by $95\%$ for each method. This significant slowdown is attributed to the high dimensionality of appearance features and the increased computational burden, especially when multi-camera data is involved. Moreover, since the primary cause of tracking failures is attributed to 3D triangulation bias, using appearance features to enhance cross-frame cross-camera person matching does not effectively address this fundamental issue.}

\revised{Additionaly, the results also demonstrate that despite the inevitable increase in computational costs from incorporating domain knowledge into our method, it sustains a competitive tracking speed, comparable to that of MvPT~\cite{dong2021fast} and UniMMT~\cite{yang2023unified}. This underscores the efficiency of our approach, even with the inclusion of more complex processes designed to enhance robustness.}

\noindent{\bf Qualitative results visualization.} 
Fig.~\ref{fig:comparation_plots} highlights the robustness of our method under challenging conditions where valid detections are limited to two opposing views. In the examples of BB, UB, and PB, UniMMT~\cite{yang2023unified} generates 3D positions with a significant bias. To alleviate this problem, our cascaded DA switches to leverage the gymnastics domain knowledge, thereby achieving more accurate 3D positions on the specified vertical plane. In the examples of SR and HB, our results exhibit minor differences to UniMMT, remaining consistent with the results in TABLE~\ref{table:compareTracking}.

We have extended the plotting tool PoseViz~\cite{isarandi_2023} to visualize estimated 3D poses. \revised{We select the two gymnastic categories most affected by tracking bias---namely Balance Beam (BB) and Uneven Bars (UB)---to demonstrate how tracking bias impacts the corresponding 3D pose estimation. This analysis is presented in Fig.~\ref{fig:pose_compare}. When only the corresponding detections in the opposite view are given, biased tracklets will misdirect pose estimation and lead to incorrect 3D poses.}

\revised{In the end, examples of pose visualizations for all gymnastic categories, obtained using our gymnast tracking method, are displayed in Figure~\ref{fig:demo}. The high-quality pose estimation resulting from the integration of gymnastics domain knowledge into our tracking method is clearly evident, further emphasizing the method's effectiveness and potential for practical application.}

\begin{table}[h!]
	\footnotesize	
	\centering
	\caption{\textbf{Ablation study for our multi-camera multi-frame DA.} \revised{The average Failure Rate and} AED values of seven types of gymnastics are used for the evaluation. The best result is rendered with \textbf{bold values}.}
	\label{table:ablation}
	\resizebox{\linewidth}{!}{
	\setlength{\tabcolsep}{.98mm}{
	\begin{tabular}{lccc}
	\toprule  
	 & \multicolumn{3}{c}{\revised{\textbf{Avg. Failure Rate $|$ AED}} }\\ 
	 \cmidrule(lr){2-4}
	 & $n_{cam}=2,$ & $~~~3,$ & $~4$\\
	 \hline
	 \rule{0pt}{4ex}   \begin{tabular}[c]{@{}l@{}} (i) Our Baseline\\ (Ray2Plane Intersection + Triangulation)\end{tabular} & \textbf{0.19\%}$|$\textbf{0.178}  & \textbf{0.08\%}$|$\textbf{0.081} & \textbf{0.07\%}$|$\textbf{0.078}\\
	 \rule{0pt}{4ex}   \begin{tabular}[c]{@{}l@{}} (ii) Remove Ray2Plane Intersection\\ (Use Triangulation Only) \end{tabular} & 1.11\%$|$0.224 & 0.12\%$|$0.085 & 0.09\%$|$0.080\\
	 \rule{0pt}{4ex}   \begin{tabular}[c]{@{}l@{}} (iii) Remove Triangulation\\ (Use Ray2Plane Intersection Only) \end{tabular} & \textbf{0.19\%}$|$\textbf{0.178} & 0.19\%$|$0.178 & 0.19\%$|$0.178\\
	\bottomrule    
\end{tabular}}}
\end{table}

\noindent{\bf Ablation study.} As indicated in TABLE~\ref{table:ablation}, our baseline, which integrates both Ray2Plane Intersection and Triangulation (\ie, case (i)), delivers the best performance for $n_{cam} = \{ 2, 3, 4\}$. By removing the Ray2Plane Intersection in our baseline, the gymnastics domain knowledge will not be utilized and the triangulation method may yield biased 3D positions due to insufficient opposite-view detections. Therefore, the overall performance of case (ii) is marginally worse than case (i). \revised{To some extent, in setting (ii), our method becomes nearly identical to UniMMT.} Nonetheless, although the target gymnast may remain on our defined central plane during the majority of performance time, whenever the target gymnast leaves the central plane, the assumption for Ray2Plane Intersection is not established. Therefore, when more than two view detections are available, switching to using triangulation generally produces better results than continuing to use Ray2Plane Intersection (\ie, case (iii)). Based on the cross-camera detections, our baseline dynamically switches the mode between Ray2Plane Intersection and Triangulation to achieve promising results.

\section{Discussion}\label{sec:discussion}

\subsection{Limitations} 
Although our method illustrates the superiority of tracking the target gymnast, certain limitations should be taken into account. 
\begin{itemize}

\item Firstly, we aim to apply our tracking framework in real-world applications, which requires a delicate balance between tracking speed and accuracy. In our pursuit of speed, our method forgoes the use of appearance features and relies solely on geometric features for tracking. Consequently, as demonstrated in our experiments, our method may still generate ID switches, necessitating additional post-processing to rectify this issue. 

\item \revised{Secondly, the unique environment of a standard gymnastics championship stadium imposes constraints on the number of cameras that can be set up for capturing. This particular situation requires us to incorporate gymnastic domain knowledge to manage insufficient opposite-view detections effectively. However, in multi-camera capturing studios where ample cameras ensure sufficient cross-view detections, our proposed method may not offer significant advantages over other SOTA multi-camera tracking methods. This suggests that the effectiveness of our method is context-dependent.}

\item \revised{Lastly, it might be challenging to distill and integrate domain knowledge into multi-camera tracking, as seen in our gymnastics case. For instance, in popular sports like soccer and basketball, there are no specific routine paths to constrain the players' movements. However, since the ground plane may constrain their movement, as their feet touch the ground most of the time, we can perform sports field registration~\cite{Gutierrez-Perez_2024_CVPR} and recover bird's-eye view (BEV) 3D tracklets from single-camera 2D tracklets. }
\end{itemize}

\subsection{Broader Impacts}

In this paper, we present a novel and robust multi-camera gymnast tracking framework designed for gymnastics judging systems. By incorporating domain-specific knowledge of gymnastics into our innovative cascaded DA, our tracking framework outperforms existing methods in the multi-camera gymnast tracking task. The impacts of our work can be summarized from two perspectives.

From the technical perspective, our framework's superior performance, especially in challenging scenarios, underscores its robustness and adaptability. This adaptability is a testament to the efficacy of incorporating domain-specific knowledge into the cascaded DA, a strategy that we believe holds promise for similar applications in multi-camera video processing.

From the social impact perspective, our work contributes to build an innovative gymnastics judge support system, which holds immense potential to revolutionize the way we watch, play, and analyze gymnastics.


\ifCLASSOPTIONcaptionsoff
  \newpage
\fi

\bibliographystyle{IEEEtran}
\bibliography{ref}

\end{document}